\documentclass{article}

\usepackage[main,final]{neurips_2026}       
\usepackage{fix-cm}

\usepackage[utf8]{inputenc}
\usepackage[T1]{fontenc}

\usepackage{hyperref}
\hypersetup{
  pdftitle={PersonaManifold: Revealing and Exploiting Curved Geometry in LLM Persona Representations},
  pdfauthor={Rui Xu, Yinghui Xu, Libo Wu}
}
\usepackage{url}
\usepackage{booktabs}
\usepackage{amsfonts}
\usepackage{nicefrac}
\usepackage{amsmath}
\usepackage{amssymb}
\usepackage{graphicx}
\usepackage{xcolor}
\usepackage{multirow}
\usepackage{enumitem}
\usepackage{subcaption}
\usepackage{wrapfig}

\title{PersonaManifold: Revealing and Exploiting\\Curved Geometry in LLM Persona Representations}

\author{
  Rui Xu$^{1,2}$ \quad Yinghui Xu$^{1}$ \quad Libo Wu$^{1,2}$ \\
  $^{1}$Fudan University \\
  $^{2}$Shanghai Innovation Institute \\
  \texttt{24110240097@m.fudan.edu.cn}
}

\begin{document}

\maketitle

\begin{abstract}
Controlling persona in large language models (LLMs) at inference time is important for role-playing, personalized dialogue, and social simulation. Recent methods extract persona vectors from the model's activation space and apply Euclidean operations---addition, scaling, and linear interpolation---under the linear representation hypothesis. However, these methods themselves report systematic failures: non-orthogonal trait dimensions, asymmetric ceiling and resistance effects, and significant deviations in multi-trait composition, suggesting that the linear isotropic assumption does not hold. We propose \textsc{PersonaManifold}, a framework that models persona representations as points on a curved, low-dimensional Riemannian submanifold in activation space. We estimate the manifold's intrinsic geometry---local metric tensors, geodesic distances, and Ollivier--Ricci curvature---and introduce \emph{geodesic steering}, which interpolates between personas along manifold geodesics rather than Euclidean straight lines. We also propose the Behavioral Similarity Triplet (BST) benchmark, which automatically generates situational questions grounded in six established psychological constructs and defines persona similarity through behavioral responses rather than self-report questionnaires. Experiments on three open-source LLMs show that persona activations form a manifold with heterogeneous curvature, geodesic distance predicts behavioral similarity more accurately than Euclidean alternatives with independent contributions from anisotropy and curvature, and geodesic steering produces more coherent intermediate personas on both our BST benchmark and external evaluations, with the advantage concentrated in high-deviation regions where the manifold deviates most from flatness.
Code is available at \url{https://github.com/airaer1998/PersonaManifold}.

\end{abstract}

\section{Introduction}

Large language models (LLMs) increasingly serve as role-playing agents, personalized assistants, and social simulation platforms, all of which require fine-grained control over the model's expressed persona \citep{mindecho2024, coser2025, consistently2025}. Among the approaches to persona control, inference-time activation steering is particularly appealing from a practical standpoint: it requires no gradient updates, enables flexible switching and composition of personality traits, and preserves the model's general capabilities. This has motivated a line of work that extracts \emph{persona vectors} from the model's activation or weight space and manipulates them through algebraic operations \citep{persona2026, persona_vectors2026, personality_vector2025}.

Several recent methods instantiate this paradigm: \citet{persona2026} extract Big Five steering directions via contrastive activation analysis, \citet{persona_vectors2026} automate the pipeline from trait descriptions to steering vectors, and \citet{personality_vector2025} compose traits in weight space via model merging. These methods share a core assumption: persona representations are \emph{linear and isotropic}---Euclidean distance and straight-line interpolation suffice for persona similarity and composition.

However, the linear isotropic assumption faces systematic challenges in practice. Existing work reports three categories of failures: (1)~\emph{Non-orthogonal dimensions}---different personality traits are not independent in activation space and exhibit significant directional coupling \citep{persona2026, five_to_many2026}; (2)~\emph{Asymmetric effects}---certain traits encounter ceiling effects or resistance during steering, indicating non-uniform constraints across directions \citep{persona2026, persona_vectors2026}; (3)~\emph{Composition deviation}---linear combination of multiple traits produces behavior that significantly deviates from the intended persona, showing that Euclidean superposition does not faithfully preserve persona semantics \citep{personality_vector2025, character_destiny2025}. These phenomena are not isolated artifacts. They jointly point to a more fundamental conclusion: the space occupied by persona representations has \emph{curved, anisotropic geometry}, and Euclidean operations cannot accurately capture its distance, directional, and compositional structure.

We propose \textsc{PersonaManifold}, a framework that models persona representations as points on a curved, low-dimensional Riemannian submanifold within the activation space. We extract persona activations from approximately 2K diverse personas across three LLMs and estimate the manifold's intrinsic geometry: local metric tensors that capture directional anisotropy, graph geodesic distances that respect the manifold's curvature, and Ollivier--Ricci curvature that quantifies local geometric heterogeneity. Based on this geometric characterization, we introduce \emph{geodesic steering}: instead of interpolating between two personas along a Euclidean straight line, we follow the manifold's geodesic path, where intermediate points remain within data-supported persona regions. We further propose the Behavioral Similarity Triplet (BST) benchmark, which automatically generates situational questions grounded in six established psychological constructs of individual behavioral differences, and defines persona similarity through behavioral responses rather than self-report questionnaires.

Experiments on multiple open-source LLMs demonstrate that geodesic distance predicts behavioral similarity more accurately than Euclidean distance, with factor separation analysis confirming independent contributions from both anisotropy and curvature. Geodesic steering produces more coherent intermediate personas than linear interpolation on both our BST benchmark and external evaluations including PersonaGym \citep{personagym2025} and the adapted BFI-44 protocol \citep{persona2026}. The advantage is concentrated in high-deviation regions where the geodesic path diverges most from the straight line, confirming that the curved geometry is practically consequential.

Our contributions are as follows:
\begin{enumerate}[leftmargin=*]
    \item We reveal that persona representations in LLMs form a curved, low-dimensional Riemannian submanifold with heterogeneous curvature. Through density-controlled analysis, we show that the curvature is an intrinsic geometric property rather than a sampling artifact.
    \item We propose geodesic steering, which interpolates between personas along manifold geodesics and produces more coherent intermediate personas than linear interpolation, validated on both internal and external benchmarks.
    \item We introduce the Behavioral Similarity Triplet (BST) benchmark, grounded in established psychological constructs, that defines persona similarity through situational behavioral responses rather than self-report questionnaires.
\end{enumerate}

\section{Related Work}

\paragraph{Persona representation and control.}
Recent inference-time steering methods extract persona vectors from LLM activations and manipulate them via algebraic operations under the linear isotropic assumption \citep{persona2026, persona_vectors2026, personality_vector2025}. However, mechanistic analyses reveal that personality traits correspond to circuits of heterogeneous sizes and depths \citep{traits_circuits2026}, opposing traits are encoded by sparse, minimally overlapping subnetworks \citep{personality_subnetworks2026}, and LLM personality structures far exceed five-dimensional complexity \citep{five_to_many2026}. These findings point to a hierarchical, anisotropic organization more naturally described by manifold geometry. Our work models persona representations as points on a curved Riemannian submanifold and shows that respecting this geometry improves both distance prediction and steering.

\paragraph{Manifold learning and Riemannian geometry.}
Classical manifold learning recovers intrinsic structure via graph shortest paths \citep{isomap2000} or diffusion processes \citep{diffusion_maps2006}, but does not characterize local curvature or metric anisotropy. Ollivier--Ricci curvature \citep{ollivier2009} extends Ricci curvature to discrete graphs, enabling geometric analyses of networks \citep{ni2019community}. In latent generative models, geodesic interpolation produces semantically smoother transitions than Euclidean interpolation \citep{rvae2019}. Within NLP, contextual embeddings exhibit anisotropy \citep{ethayarajh2019} and concentrate in narrow cones \citep{cai2021isotropy}, yet these analyses remain at the token level. The Riemannian geometry of \emph{persona} representations has not been characterized, nor exploited for activation steering.

\section{Method}

\begin{figure}[t]
    \centering
    \includegraphics[width=\textwidth]{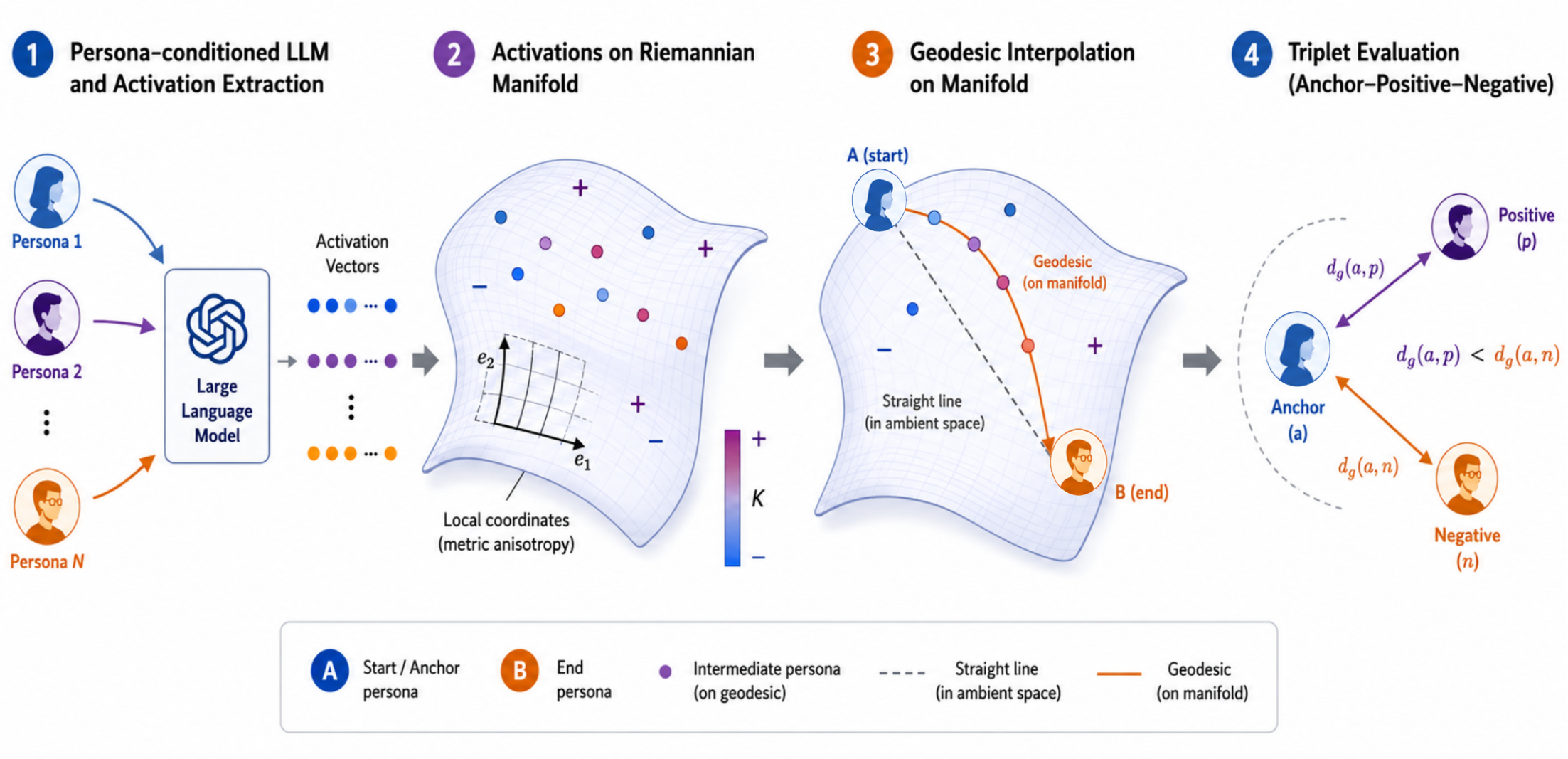}
    \caption{Overview of \textsc{PersonaManifold}. (1) Extract persona activations from an LLM's residual stream. (2) Estimate manifold geometry: local metric tensors, geodesic distances, and Ollivier--Ricci curvature. (3) Geodesic steering interpolates along the manifold surface. (4) BST benchmark evaluates behavioral similarity via triplets.}
    \label{fig:main}
\end{figure}

We present an overview of \textsc{PersonaManifold} in Figure~\ref{fig:main}. The pipeline proceeds in four stages: extracting persona activations (\S\ref{sec:extraction}), estimating manifold geometry (\S\ref{sec:geometry}), geodesic steering (\S\ref{sec:steering}), and BST evaluation (\S\ref{sec:bst}).

\subsection{Persona Activation Extraction}
\label{sec:extraction}

We construct a persona pool $\mathcal{P} = \{p_1, \ldots, p_N\}$ of $N \approx 2\text{K}$ personas, where each persona $p_i$ is a textual description used as the system prompt to condition the LLM (e.g., ``a conflict-averse pediatrician who prioritizes harmony''). Personas are drawn from two complementary sources: (1) a stratified subset of 1.5K personas from PersonaHub \citep{personahub2024, capturing_minds2024}, covering professions, demographics, and value profiles; and (2) 500 literary characters from CoSER \citep{coser2025, character_understanding2024}, prioritizing rare trait combinations underrepresented in synthetic data. This combination ensures both breadth across common persona types and coverage of the manifold's sparse, high-curvature regions.

Each persona $p_i$ is paired with $M = 40$ probe questions designed to elicit trait-relevant behavioral variation. We adapt items from BFI-44 \citep{john1991bfi}, HEXACO-60 \citep{ashton2009hexaco}, and TIPI \citep{gosling2003tipi}, rephrasing each into an open-ended situational prompt using GPT-5.2 and manually reviewing for clarity (Appendix~\ref{app:probes}). For each pair $(p_i, q_j)$, we extract the residual stream activation $h^{(\ell)}_{i,j}$ at the last token of the model's generated response from a single layer $\ell$. The persona representation is the mean activation with a neutral baseline subtracted:
\begin{equation}
    x_i = \frac{1}{M} \sum_{j=1}^{M} \big(h^{(\ell)}_{i,j} - h^{(\ell)}_{0,j}\big) \in \mathbb{R}^D,
\end{equation}
where $h^{(\ell)}_{0,j}$ is the activation from a neutral (no-persona) prompt. Subtracting the baseline ensures that $x_i$ encodes the persona's deviation from the neutral state. We select a single layer $\ell$ per model from candidates in the range 12--20, where persona information concentrates \citep{traits_circuits2026, dissecting_persona2025}. Specifically, we compute validation TCR on a held-out set of 200 personas for each candidate layer and select the one with the highest TCR; for all three models, the optimal layer falls in the 12--16 range (see Appendix~\ref{app:layer} for the full layer sweep). We then apply PCA retaining 99\% cumulative variance ($D \approx 4096 \to D' \in [200, 500]$) to remove noise dimensions while preserving curved structure.

\subsection{Manifold Geometry Estimation}
\label{sec:geometry}

We characterize the persona manifold along three axes: its intrinsic dimensionality, the directional dependence of local distances, and its curvature. These are addressed through intrinsic dimensionality estimation, local metric tensors, geodesic distances, and curvature analysis, respectively.

\paragraph{Intrinsic dimensionality.}
We estimate $d^*$ using two complementary methods---the Marchenko--Pastur spectral test on local covariances and the MLE of \citet{levina2004}---and take their consensus.

\paragraph{Local metric tensor.}
Euclidean distance implicitly assumes all directions are equally important. In persona space, some directions vary frequently (e.g., extroversion--introversion) while others are rare (e.g., unusual value combinations). A unit displacement along a rare direction should count as ``farther.'' For each $x_i$, we compute the local covariance over its $k$-nearest neighbors ($k \geq 3d^*$):
\begin{equation}
    C_i = \frac{1}{k} \sum_{x_j \in \mathcal{N}_k(x_i)} (x_j - x_i)(x_j - x_i)^\top.
\end{equation}
Let $\{v_i^{(a)}, \lambda_a(x_i)\}_{a=1}^{d^*}$ be the top $d^*$ eigenpairs. The metric tensor is:
\begin{equation}
    g(x_i) = \sum_{a=1}^{d^*} \frac{1}{\lambda_a(x_i) + \epsilon}\, v_i^{(a)} {v_i^{(a)}}^\top,
\end{equation}
where $\epsilon$ is chosen by cross-validation. Large $\lambda_a$ means the direction is ``cheap'' (common variation); small $\lambda_a$ means ``expensive'' (rare variation).

\paragraph{Geodesic distance.}
We construct a $k$-NN graph with edge weights that respect the local metric:
\begin{equation}
    w_{ij} = \tfrac{1}{2}\big(\|x_j - x_i\|_{g(x_i)} + \|x_i - x_j\|_{g(x_j)}\big).
\end{equation}
The geodesic distance $d_g(x_i, x_j)$ is the Dijkstra shortest path length. It captures two effects: (a) the anisotropic metric biases paths toward low-cost directions; (b) the graph topology forces paths to follow data-dense regions rather than cutting through empty space. We validate on synthetic manifolds with known geodesics and report bootstrap stability (Kendall's $\tau$ under 10\% subsampling).

\paragraph{Curvature analysis.}
We adopt Ollivier--Ricci curvature \citep{ollivier2009}:
\begin{equation}
    \mathrm{Ric}(x_i, x_j) = 1 - \frac{W_1(\mu_i, \mu_j)}{d_g(x_i, x_j)},
\end{equation}
where $\mu_i$ is the normalized neighbor measure and $W_1$ is the Wasserstein-1 distance. Positive curvature indicates local convergence (like a sphere); negative curvature indicates divergence (like a saddle). Since $\mathrm{Ric}$ is defined on edges, we assign each node $x_i$ a scalar curvature by averaging over its $k$-NN edges: $\kappa(x_i) = \frac{1}{k}\sum_{x_j \in \mathcal{N}_k(x_i)} \mathrm{Ric}(x_i, x_j)$. To distinguish intrinsic geometry from sampling artifacts, we perform \emph{density--curvature decoupling}: we bin points by local density and verify that curvature differences persist within density-matched subsets.

\subsection{Geodesic Activation Steering}
\label{sec:steering}

Given two target personas $p_1, p_2$ with representations $x_1, x_2$, linear steering interpolates along the straight line $v(t) = (1-t)x_1 + tx_2$, whose midpoints may lie outside the manifold in low-density regions. Geodesic steering instead follows the manifold surface, keeping intermediates within data-supported persona regions.

\paragraph{Procedure.}
\begin{enumerate}[leftmargin=*]
    \item Compute the shortest path $\{x_1 = z_0, z_1, \ldots, z_L = x_2\}$ on the $k$-NN graph, smooth via cubic spline in local tangent spaces to obtain $\gamma: [0, 1] \to \mathbb{R}^{D'}$.
    \item Sample $K$ intermediate points $\{\gamma(t_1), \ldots, \gamma(t_K)\}$ at equal arc-length intervals.
    \item Project each intermediate back to the original activation space via the inverse PCA mapping $P^\dagger: \mathbb{R}^{D'} \to \mathbb{R}^D$, and inject into the residual stream:
\end{enumerate}
\begin{equation}
    h_\ell \leftarrow h_\ell + \alpha \cdot \frac{P^\dagger \gamma(t_k)}{\|P^\dagger \gamma(t_k)\|} \cdot s(t_k),
\end{equation}
where $s(t_k) = (1-t_k)\|x_1\| + t_k\|x_2\|$ ensures smooth steering intensity between endpoints. Because $P^\dagger$ preserves norms on the PCA subspace ($\|P^\dagger z\| = \|z\|$ for $z \in \mathbb{R}^{D'}$), the PCA-space norms $\|x_1\|, \|x_2\|$ are consistent with the original activation scale.

The advantage over linear steering is largest when $d_g/d_E$ is high---i.e., the geodesic deviates substantially from the straight line. When the manifold is locally flat ($d_g/d_E \approx 1$), the two methods are equivalent.

\subsection{Behavioral Similarity Triplet (BST)}
\label{sec:bst}

Evaluating persona similarity through self-report questionnaires is problematic: recent work has shown that LLMs' self-reported personality traits systematically dissociate from their actual behavior \citep{personality_illusion2026, incharacter2024}. We therefore propose BST, a benchmark that defines persona similarity through observed behavioral responses to situational questions. The construction proceeds in three stages.

\paragraph{Question generation.}
We select six established psychological constructs---Moral Foundations Theory \citep{graham2013mft}, DOSPERT \citep{blais2006dospert}, the Interpersonal Circumplex \citep{wiggins1979circumplex}, Decision-Making Style \citep{scott1995decision}, Schwartz Values \citep{schwartz1992values}, and Communicator Style \citep{norton1978communicator}---and generate 10 forced-choice and 20 open-ended candidate questions per construct using GPT-5.2 (180 total; see Appendix~\ref{app:bst_prompts}). After discriminativeness filtering (removing questions with response entropy $< 0.8$ nats or embedding variance $< 0.05$; see Appendix~\ref{app:bst_prompts} for details), we retain 150 questions (48 forced-choice + 102 open-ended).

\paragraph{Behavioral distance and triplets.}
We define $d_\text{behav} = \beta \cdot d_\text{choice} + (1-\beta) \cdot d_\text{emb}$, where $d_\text{choice}$ is option agreement on forced-choice questions, $d_\text{emb}$ is embedding cosine on open-ended responses, and $\beta = 1/3$. We set $\beta < 0.5$ because the 102 open-ended questions carry richer behavioral signal than the 48 forced-choice items (higher per-question information content), though we report TCR-choice separately to ensure conclusions do not depend on this weighting. For each anchor, we select positives from the nearest 5\% and negatives from the farthest 20\% (requiring $d(a,n) \geq 3 \cdot d(a,p)$), yielding 10K triplets with $\geq$85\% human agreement (Appendix~\ref{app:human_agreement}).

\paragraph{Evaluation metric: TCR.}
$\text{TCR} = \Pr[d(a,p) < d(a,n)]$ over BST triplets. We report TCR-choice (purely discrete), TCR-emb, and TCR-combined. TCR-choice rules out circularity when evaluating embedding-based metrics.

\section{Experiments}

We organize experiments around three research questions: \textbf{RQ1}---Do persona activations form a curved manifold? \textbf{RQ2}---Does geodesic distance predict behavioral similarity better than Euclidean alternatives? \textbf{RQ3}---Does geodesic steering produce more coherent intermediate personas?

\subsection{Experimental Setup}
\label{sec:setup}

\paragraph{Models.}
We extract persona activations from three open-source LLMs: Llama-3.1-8B-Instruct, Qwen2.5-7B-Instruct, and Mistral-7B-Instruct-v0.3. For each model, we select a single optimal layer from the 12--16 range via validation TCR (layer 16 for Llama, 16 for Qwen, 14 for Mistral; see Appendix~\ref{app:layer}). These models are chosen to match the experimental settings of prior persona steering work \citep{persona2026, personality_vector2025} and to span three distinct pre-training pipelines at comparable scale ($\sim$7--8B parameters), enabling architecture-controlled comparison.

\paragraph{Persona set.}
We use the same persona pool described in \S\ref{sec:extraction}: $\sim$2K personas from PersonaHub \citep{personahub2024} and CoSER \citep{coser2025}, each paired with the 40 probe questions from \S\ref{sec:extraction}.

\paragraph{Distance baselines.}
We compare seven distance metrics organized along a factor separation gradient: \emph{flat isotropic} (Euclidean, Cosine), \emph{flat anisotropic} (Mahalanobis), \emph{curved isotropic} (Isomap-Euclidean, Diffusion Map), \emph{nonlinear embedding} (UMAP), and \emph{curved anisotropic} (our Geodesic). This gradient isolates each geometric factor: Euclidean $\to$ Mahalanobis tests anisotropy; Mahalanobis $\to$ Isomap tests curvature; Isomap $\to$ Geodesic tests their interaction.

\paragraph{Steering baselines.}
For steering, we compare: Euclidean-Linear (default in \citealt{persona2026} and \citealt{persona_vectors2026}), SLERP, Isomap-Euclidean, Diffusion Map, UMAP, Graph-NN Chain (graph path without spline smoothing), and our Geodesic Steering.

\paragraph{Evaluation.}
Behavioral similarity is evaluated via TCR on the BST benchmark (\S\ref{sec:bst}). Steering quality is assessed by three internal metrics. \emph{Interpolation Coherence} (IC) is the mean pairwise cosine similarity of response embeddings (via \texttt{all-MiniLM-L6-v2}) across $N{=}20$ dialogue turns at each intermediate point, averaged over all $K$ intermediates; higher IC indicates more self-consistent behavior. \emph{Transition Smoothness} (TS) measures the fraction of (trait, step) pairs where BFI-44 trait scores progress monotonically toward the target: $\text{TS} = \frac{1}{5K}\sum_{d,k} \mathbf{1}[\text{sgn}(\Delta s_{d,k}) {=} \text{sgn}(s_d(1) {-} s_d(0))]$, where $s_d(t)$ is the $d$-th Big Five score at interpolation point $t$ (obtained via the BFI-44 protocol of \citealt{persona2026}). \emph{Manifold Adherence} (MA) quantifies how close intermediates stay to observed personas: $\text{MA} = 1 - \frac{1}{K}\sum_k d_g(\gamma(t_k), \text{NN}(\gamma(t_k))) / r_{95}$, where $\text{NN}(\cdot)$ returns the nearest observed persona and $r_{95}$ is the 95th-percentile inter-persona geodesic distance. We also evaluate on external benchmarks: PersonaGym PersonaScore \citep{personagym2025}, BFI-44 Monotonicity (adapted from \citealt{persona2026}), and Trait Expression Score \citep{persona_vectors2026}, as well as human naturalness ratings.

\subsection{Persona Activations Form a Curved Manifold (RQ1)}
\label{sec:exp_manifold}

\begin{table}[t]
\centering
\caption{Intrinsic dimensionality estimates $d^*$ across models and layers. MP: Marchenko--Pastur; LB: Levina--Bickel MLE. Both methods agree within $\pm 2$ dimensions.}
\label{tab:dimensionality}
\small
\begin{tabular}{llccc}
\toprule
\textbf{Model} & \textbf{Method} & \textbf{Layer 12} & \textbf{Layer 16} & \textbf{Layer 20} \\
\midrule
\multirow{2}{*}{Llama-3.1-8B}  & MP & 22 & 23 & 21 \\
                                & LB & 21 & 22 & 20 \\
\midrule
\multirow{2}{*}{Qwen2.5-7B}    & MP & 19 & 20 & 18 \\
                                & LB & 18 & 19 & 17 \\
\midrule
\multirow{2}{*}{Mistral-7B}    & MP & 16 & 17 & 15 \\
                                & LB & 15 & 16 & 15 \\
\bottomrule
\end{tabular}
\end{table}

As reported in Table~\ref{tab:dimensionality}, both Marchenko--Pastur and Levina--Bickel estimators yield $d^* \in [15, 23]$ across all three models, far below the ambient dimension ($D' \approx 300$) yet substantially above the 5 assumed by Big Five models. This confirms the existence of a low-dimensional manifold whose complexity exceeds traditional psychological frameworks. The two estimators agree across models and layers, indicating that $d^*$ is robust.

\begin{figure}[t]
    \centering
    \includegraphics[width=\textwidth]{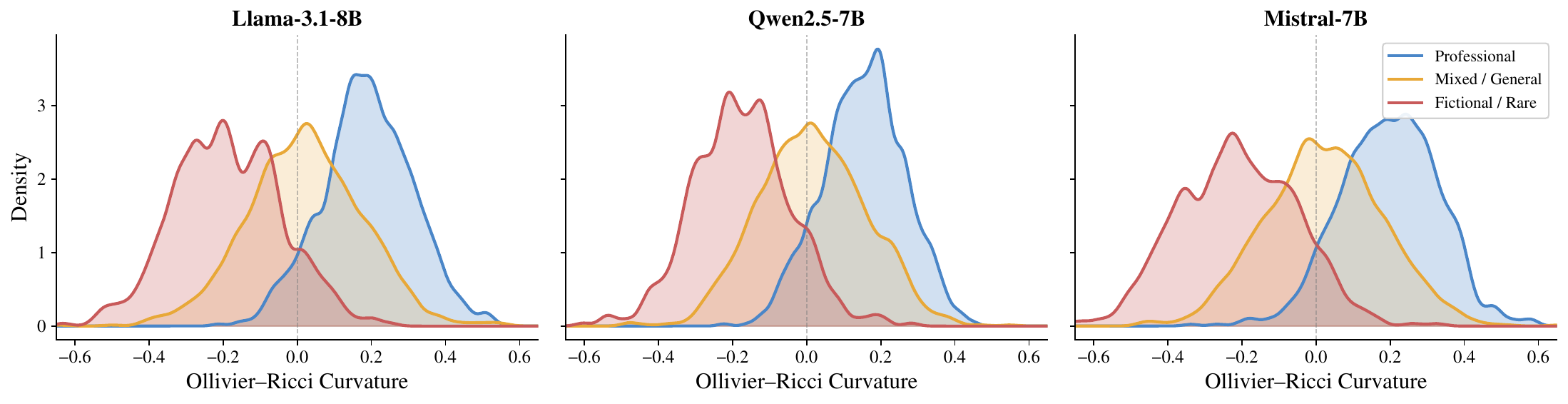}
    \caption{Ollivier--Ricci curvature distribution by persona type. Professional personas tend toward positive curvature; fictional characters exhibit negative curvature. Consistent across architectures.}
    \label{fig:curvature}
\end{figure}

As shown in Figure~\ref{fig:curvature}, Ollivier--Ricci curvature varies systematically across persona types. Professional personas (teachers, doctors, engineers) tend toward positive curvature, forming dense clusters of similar roles. In contrast, fictional characters and rare trait combinations exhibit negative curvature, occupying isolated neighborhoods that diverge from one another. A density--curvature decoupling analysis (Appendix~\ref{app:decoupling}) confirms that this heterogeneity reflects intrinsic geometry rather than non-uniform sampling: within density-matched subsets, curvature differences across persona types remain significant ($p < 0.001$).

\subsection{Geodesic Distance Predicts Behavioral Similarity (RQ2)}
\label{sec:exp_bst}

\begin{table}[t]
\centering
\caption{Triplet Consistency Rate (\%) on BST. Best in \textbf{bold}. $\Delta$: improvement over Euclidean. All significant at $p < 0.01$ (bootstrap).}
\label{tab:tcr}
\small
\setlength{\tabcolsep}{4pt}
\begin{tabular}{l ccc ccc ccc}
\toprule
& \multicolumn{3}{c}{\textbf{Llama-3.1-8B}} & \multicolumn{3}{c}{\textbf{Qwen2.5-7B}} & \multicolumn{3}{c}{\textbf{Mistral-7B}} \\
\cmidrule(lr){2-4} \cmidrule(lr){5-7} \cmidrule(lr){8-10}
\textbf{Metric} & Choice & Emb & Comb & Choice & Emb & Comb & Choice & Emb & Comb \\
\midrule
Euclidean       & 60.2 & 62.8 & 62.1 & 59.5 & 61.3 & 61.5 & 58.8 & 60.7 & 60.8 \\
Cosine          & 59.8 & 62.1 & 61.5 & 59.1 & 60.9 & 61.0 & 58.3 & 60.2 & 60.1 \\
Mahalanobis     & 62.5 & 65.1 & 64.3 & 61.8 & 63.7 & 63.9 & 61.0 & 63.1 & 63.0 \\
Isomap-Euclid.  & 63.8 & 66.2 & 65.8 & 63.0 & 64.9 & 65.1 & 62.5 & 64.5 & 64.3 \\
Diffusion Map   & 63.1 & 65.5 & 65.0 & 62.4 & 64.7 & 64.5 & 61.8 & 63.8 & 63.7 \\
UMAP            & 62.3 & 64.8 & 64.2 & 61.5 & 63.5 & 63.6 & 61.5 & 63.3 & 63.1 \\
\textbf{Geodesic (ours)} & \textbf{65.2} & \textbf{68.4} & \textbf{68.2} & \textbf{64.3} & \textbf{66.8} & \textbf{67.0} & \textbf{63.9} & \textbf{66.1} & \textbf{65.9} \\
\midrule
$\Delta$ vs.\ Euclid. & +5.0 & +5.6 & +6.1 & +4.8 & +5.5 & +5.5 & +5.1 & +5.4 & +5.1 \\
\bottomrule
\end{tabular}
\end{table}

Results in Table~\ref{tab:tcr} show that geodesic distance consistently achieves the highest TCR across all three splits and all three models, with absolute improvements of +5.1--6.1 percentage points in TCR-combined over Euclidean distance (bootstrap 95\% CI excludes zero). Notably, the advantage holds on TCR-choice, which is computed entirely from discrete option agreement without any embedding, ruling out the circularity of validating one Euclidean metric with another.

\begin{figure}[t]
    \centering
    \includegraphics[width=\textwidth]{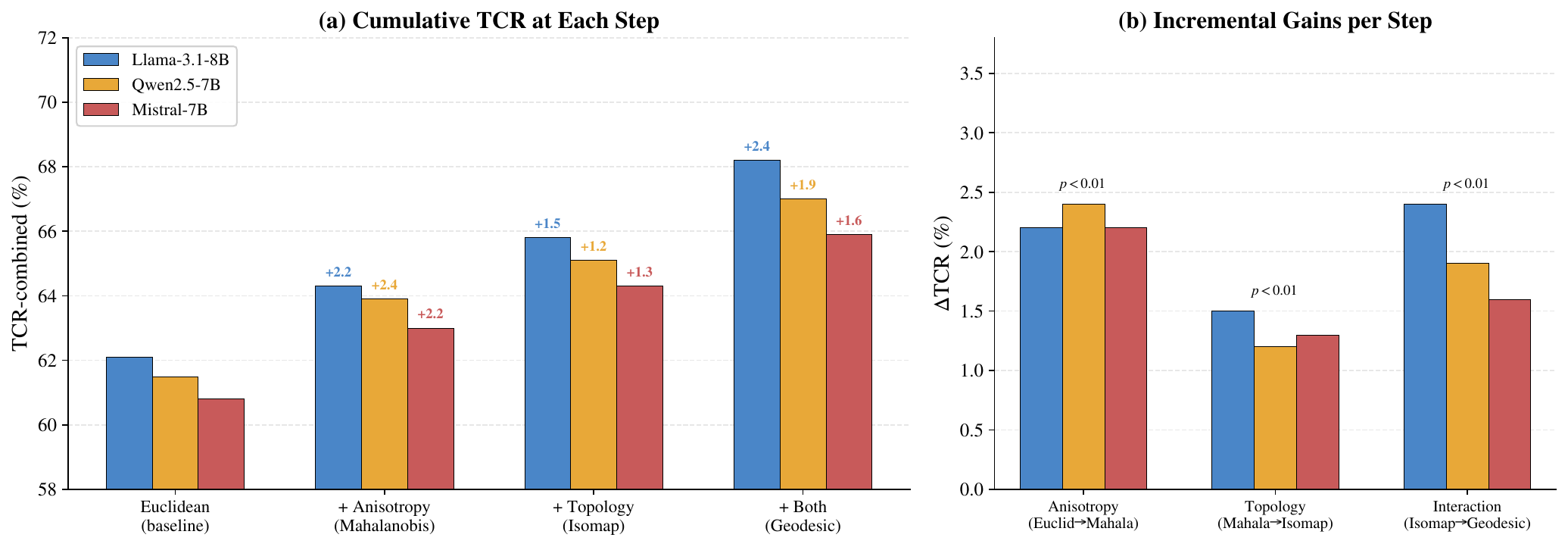}
    \caption{Factor separation waterfall plot. TCR-combined improves monotonically at each step: Euclidean $\to$ Mahalanobis (anisotropy), $\to$ Isomap (graph topology), $\to$ Geodesic (both). Each step is significant ($p < 0.01$, bootstrap).}
    \label{fig:factor_separation}
\end{figure}

To understand where the improvement originates, Figure~\ref{fig:factor_separation} decomposes TCR gains along the factor separation gradient. Introducing anisotropy alone (Euclidean $\to$ Mahalanobis) yields +2--2.5\%; adding graph topology (Mahalanobis $\to$ Isomap) contributes +1--1.5\%; combining both in the full geodesic adds +1.5--2.5\%. All three steps are statistically significant ($p < 0.01$, bootstrap), confirming that the two geometric properties provide complementary information. The relative importance varies by persona type: for professional personas (positive curvature), the anisotropy step contributes most; for fictional characters (negative curvature), the topology step dominates.

Stratifying persona pairs by $d_g/d_E$ quartile reveals that the top quartile (where the geodesic deviates most from the straight line) shows +11--12\% TCR improvement, while the bottom quartile (near-flat manifold) shows only +1--1.5\%. The practical impact of curved geometry is non-uniform---our method benefits most where the linear assumption fails hardest.

\subsection{Geodesic Steering Produces More Coherent Intermediates (RQ3)}
\label{sec:exp_steering}

We select 200 persona pairs for steering evaluation: 50 high-deviation (top-25\% by $d_g/d_E$), 50 low-deviation (bottom-25\%), and 100 random. For each pair, we sample 10 intermediate points along each method's path and generate at least 20 dialogue turns per point.

\begin{table}[t]
\centering
\caption{Steering quality on high-deviation pairs (top-25\% by $d_g/d_E$). IC: Interpolation Coherence, TS: Transition Smoothness, MA: Manifold Adherence. Best in \textbf{bold}.}
\label{tab:steering}
\small
\setlength{\tabcolsep}{3.5pt}
\begin{tabular}{l ccc ccc ccc}
\toprule
& \multicolumn{3}{c}{\textbf{Llama-3.1-8B}} & \multicolumn{3}{c}{\textbf{Qwen2.5-7B}} & \multicolumn{3}{c}{\textbf{Mistral-7B}} \\
\cmidrule(lr){2-4} \cmidrule(lr){5-7} \cmidrule(lr){8-10}
\textbf{Method} & IC & TS & MA & IC & TS & MA & IC & TS & MA \\
\midrule
Euclidean-Linear & 0.58 & 0.52 & 0.41 & 0.55 & 0.49 & 0.38 & 0.53 & 0.47 & 0.36 \\
SLERP            & 0.60 & 0.55 & 0.43 & 0.58 & 0.53 & 0.41 & 0.56 & 0.51 & 0.39 \\
Isomap-Euclid.   & 0.65 & 0.61 & 0.55 & 0.62 & 0.57 & 0.52 & 0.60 & 0.56 & 0.49 \\
Diffusion Map    & 0.63 & 0.59 & 0.52 & 0.60 & 0.55 & 0.49 & 0.58 & 0.54 & 0.47 \\
UMAP             & 0.61 & 0.56 & 0.48 & 0.58 & 0.53 & 0.45 & 0.57 & 0.51 & 0.43 \\
Graph-NN Chain   & 0.67 & 0.58 & 0.57 & 0.65 & 0.54 & 0.54 & \textbf{0.66} & 0.53 & \textbf{0.56} \\
\textbf{Geodesic (ours)} & \textbf{0.72} & \textbf{0.69} & \textbf{0.62} & \textbf{0.69} & \textbf{0.65} & \textbf{0.58} & \textbf{0.66} & \textbf{0.63} & 0.55 \\
\bottomrule
\end{tabular}
\end{table}

As shown in Table~\ref{tab:steering}, geodesic steering achieves the best IC and TS on high-deviation pairs across all three models, and the best MA on two of three (Graph-NN Chain slightly exceeds geodesic on Mistral MA). On low-deviation pairs, all methods perform comparably---as expected, since the geodesic approximately coincides with the straight line when the manifold is locally flat.

External benchmarks corroborate the internal metrics (full results in Appendix~\ref{app:external}): geodesic intermediates score higher on PersonaGym PersonaScore (1--5), exhibit smoother BFI-44 trait transitions along the interpolation path, and achieve continuously progressing Trait Expression Scores (0--100) where linear paths plateau or oscillate. One minor exception is Mistral-7B, where SLERP slightly outperforms geodesic on TS for low-deviation pairs (0.64 vs.\ 0.63). Human evaluation (three annotators, 100 dialogues, 1--5 Likert) confirms the pattern: geodesic steering receives a mean naturalness score of 3.82 vs.\ 3.41 for linear (Krippendorff's $\alpha = 0.73$).

We also apply geodesic distance to few-shot persona retrieval (Appendix~\ref{app:retrieval}), where it improves MRR on rare personas but offers little advantage on common ones where Mahalanobis suffices.

\paragraph{Cross-model consistency.}
The qualitative findings replicate across all three architectures: $d^*$ falls in the 15--23 range, curvature is heterogeneous and survives density decoupling, geodesic distance achieves higher TCR, and the steering advantage concentrates in high-deviation regions. Quantitative values differ (e.g., $d^* = 22\text{--}23$ for Llama vs.\ $16\text{--}17$ for Mistral), but method rankings are consistent, suggesting that manifold structure is a general property of LLM persona representations. As a practical guideline, geodesic steering is most beneficial when $d_g/d_E > 1.2$, which covers approximately 60\% of random persona pairs.

\section{Analysis}

\subsection{Ablation Studies}

\begin{table}[t]
\centering
\caption{Ablation studies. TCR: TCR-combined (\%). IC/TS: Llama high-deviation pairs. Default marked with $\dagger$.}
\label{tab:ablation}
\small
\setlength{\tabcolsep}{3.5pt}
\begin{tabular}{ll ccc cc}
\toprule
& & \multicolumn{3}{c}{\textbf{TCR-combined (\%)}} & \multicolumn{2}{c}{\textbf{Steering (Llama)}} \\
\cmidrule(lr){3-5} \cmidrule(lr){6-7}
\textbf{Component} & \textbf{Variant} & Llama & Qwen & Mistral & IC & TS \\
\midrule
\multirow{3}{*}{Metric est.}
    & Local PCA$^\dagger$      & 68.2 & 67.0 & 65.9 & 0.72 & 0.69 \\
    & Diffusion kernel         & 67.5 & 66.8 & 65.5 & 0.72 & 0.68 \\
    & Isotropic Euclidean      & 63.8 & 63.1 & 62.0 & 0.65 & 0.61 \\
\midrule
\multirow{3}{*}{Graph $k$}
    & $k = 3d^*$               & 66.9 & 65.7 & 64.5 & 0.70 & 0.67 \\
    & $k = 5d^*$$^\dagger$     & 68.2 & 67.0 & 65.9 & 0.72 & 0.69 \\
    & $k = 10d^*$              & 67.4 & 66.8 & 65.6 & 0.71 & 0.69 \\
\midrule
\multirow{3}{*}{Layer}
    & Shallow (4--8)           & 63.1 & 62.3 & 61.5 & 0.63 & 0.58 \\
    & Middle (12--16)$^\dagger$& 68.2 & 67.0 & 65.9 & 0.72 & 0.69 \\
    & Deep (20--24)            & 65.8 & 65.2 & 64.6 & 0.68 & 0.66 \\
\midrule
\multirow{3}{*}{PCA var.}
    & 95\%                     & 67.1 & 66.3 & 65.4 & 0.71 & 0.68 \\
    & 99\%$^\dagger$           & 68.2 & 67.0 & 65.9 & 0.72 & 0.69 \\
    & 99.5\%                   & 68.0 & 66.9 & 65.8 & 0.72 & 0.69 \\
\midrule
\multirow{2}{*}{Steering $\alpha$}
    & Fixed $\alpha$           & \multicolumn{3}{c}{---} & 0.69 & 0.65 \\
    & Norm-aware$^\dagger$     & \multicolumn{3}{c}{---} & 0.72 & 0.69 \\
\midrule
\multirow{2}{*}{Smoothing}
    & No spline (Graph-NN)     & \multicolumn{3}{c}{---} & 0.67 & 0.58 \\
    & Cubic spline$^\dagger$   & \multicolumn{3}{c}{---} & 0.72 & 0.69 \\
\bottomrule
\end{tabular}
\end{table}

Table~\ref{tab:ablation} reports ablations across all three models. Local PCA achieves the best TCR; diffusion kernel performs comparably; isotropic Euclidean degrades substantially, confirming that anisotropy is essential. TCR is stable across $k \in \{3d^*, 5d^*, 10d^*\}$: the largest gap from the default $k{=}5d^*$ is 1.3\% (Llama at $k{=}3d^*$) and the smallest is 0.3\% (Mistral at $k{=}10d^*$). Middle layers (12--16) yield the strongest manifold structure, consistent with \citet{traits_circuits2026} and \citet{dissecting_persona2025}. PCA variance threshold has minimal impact. For steering (Llama), norm-aware $\alpha$ improves IC and TS by 3--4 points on high-deviation pairs; spline smoothing improves TS without sacrificing IC or MA.

\subsection{Geometry--Performance Interaction across Persona Types}

The relative contribution of anisotropy and curvature depends on persona type. For professional personas (positive curvature, dense clusters), the Mahalanobis step accounts for the majority of the TCR gain over Euclidean---direction-aware scaling is the primary bottleneck when neighbors are plentiful but unevenly distributed. For fictional characters and rare trait combinations (negative curvature, sparse regions), the Isomap step contributes more---these personas benefit from path-following that avoids cutting through empty space. This interaction explains why the full geodesic, which combines both factors, consistently outperforms either alone, and why the advantage is largest in high-deviation regions where both anisotropy and curvature are pronounced.

\section{Conclusion}

We present PersonaManifold, a framework that models LLM persona representations as points on a curved Riemannian submanifold. Geodesic distance predicts behavioral similarity more accurately than Euclidean alternatives, and geodesic steering produces more coherent intermediate personas---with the advantage concentrated where the manifold deviates most from flatness. These findings, validated on our behaviorally grounded BST benchmark and multiple external evaluations, suggest that respecting the intrinsic geometry of internal representations is a practically useful inductive bias for persona control.

\bibliographystyle{plainnat}
\bibliography{custom}

\newpage
\appendix

\numberwithin{table}{section}
\numberwithin{figure}{section}

\section*{Limitations and Broader Impact}

Our experiments focus on 7--8B parameter models; extending the geometric analysis to larger scales (70B+) is a natural next step. The full geometry estimation ($k$-NN graph, metric tensors, curvature) requires $O(N^2)$ pairwise computations, which could be addressed with approximate nearest-neighbor methods for larger persona sets. Each persona is currently represented as a single mean activation vector, collapsing intra-persona variation across conversational contexts. Finally, the BST benchmark covers six psychological constructs; incorporating additional dimensions such as cognitive style or emotional regulation would broaden its coverage.

Improved persona control can benefit personalized dialogue systems, social simulation research, and educational role-playing applications. However, more coherent persona steering could also lower the barrier for generating convincing impersonation or social engineering content. We note that our method operates on existing open-source LLMs and does not introduce fundamentally new generative capabilities; the risk is incremental rather than qualitative. Nonetheless, we encourage practitioners to pair fine-grained persona control with robust content-safety filters and to restrict deployment contexts where impersonation poses direct harm.

\section{Implementation Details}
\label{app:implementation}

\paragraph{Compute resources.}
All experiments are conducted on a single node with 4$\times$ NVIDIA A100 80GB GPUs. Persona activation extraction for $\sim$2K personas takes approximately 45 minutes per model using batch size 16. Manifold geometry estimation ($k$-NN graph construction, local PCA, Dijkstra shortest paths, Ollivier--Ricci curvature) takes $\sim$15 minutes per model on CPU. Steering generation for the full evaluation set (200 pairs $\times$ 10 intermediates $\times$ 20 turns) takes 6--8 hours per model.

\paragraph{Software.}
We use PyTorch 2.1 and the HuggingFace Transformers library for activation extraction. Manifold geometry estimation uses scikit-learn (PCA, $k$-NN), NetworkX (graph construction, Dijkstra), and POT \citep{flamary2021pot} (optimal transport for Ollivier--Ricci curvature). BST question generation and persona dialogue generation use the OpenAI API (GPT-5.2).

\paragraph{Hyperparameters.}
Table~\ref{tab:hyperparams} reports all hyperparameters. Values were selected via grid search on a held-out validation set of 200 personas (10\% of the full pool).

\begin{table}[h]
\centering
\caption{Hyperparameters for \textsc{PersonaManifold}.}
\label{tab:hyperparams}
\small
\setlength{\tabcolsep}{4pt}
\begin{tabular}{llc}
\toprule
\textbf{Component} & \textbf{Parameter} & \textbf{Value} \\
\midrule
\multirow{4}{*}{Extraction}
    & Probe questions per persona & 40 \\
    & Target layers & 12--16 \\
    & PCA variance threshold & 99\% \\
    & Ambient \texorpdfstring{$\to$}{to} reduced dim ($D \to D'$) & 4096 \texorpdfstring{$\to$}{to} 200--500 \\
\midrule
\multirow{4}{*}{Geometry}
    & $k$-NN graph $k$ & $5d^*$ \\
    & Metric regularization $\epsilon$ & $10^{-4}$ \\
    & Curvature neighbor measure & uniform over $k$-NN \\
    & Density bins (decoupling) & 5 quintiles \\
\midrule
\multirow{4}{*}{Steering}
    & Intermediate points $K$ & 10 \\
    & Spline type & cubic \\
    & Norm-aware $\alpha$ range & [0.5, 2.0] \\
    & Dialogue turns per point & 20 \\
\midrule
\multirow{3}{*}{BST}
    & Forced-choice questions & 48 \\
    & Open-ended questions & 102 \\
    & $\beta$ (choice weight) & 1/3 \\
\bottomrule
\end{tabular}
\end{table}

\paragraph{Computational cost comparison.}
The geodesic distance computation adds overhead relative to Euclidean distance. For $N = 2\text{K}$ personas, constructing the $k$-NN graph takes $O(Nk)$ with approximate nearest neighbors (FAISS), and Dijkstra shortest paths for all pairs takes $O(N^2 \log N)$. In practice, the full geometry estimation pipeline adds $\sim$15 minutes on top of the 45-minute extraction step, a $\sim$33\% overhead that is paid once per model and amortized across all downstream evaluations. At inference time, geodesic steering requires a single graph shortest-path query per persona pair ($<$1ms), making it negligible compared to the LLM generation cost.

\section{Probe Questions}
\label{app:probes}

We design 40 probe questions to elicit trait-relevant behavioral variation across personas. Items are adapted from three established personality inventories---BFI-44 \citep{john1991bfi}, HEXACO-60 \citep{ashton2009hexaco}, and TIPI \citep{gosling2003tipi}---by rephrasing each into an open-ended situational prompt. The rephrasing was performed using GPT-5.2 with manual review for clarity and diversity.

\paragraph{Design principles.}
(1)~Each question targets a specific personality facet (e.g., agreeableness--compliance, openness--aesthetics).
(2)~Questions describe concrete situations requiring a behavioral response, not abstract self-assessments.
(3)~The set covers all six HEXACO domains to ensure breadth.
(4)~Pilot testing on 200 personas confirmed that all 40 questions produce non-degenerate response distributions (entropy $> 1.0$ nats for forced-choice, embedding variance $> 0.05$ for open-ended).

\paragraph{Rephrasing prompt.}
The following prompt is used to rephrase each inventory item into a situational behavioral question:

\begin{table}[h]
\centering
\caption{Prompt template used to rephrase personality inventory items into situational behavioral probes.}
\label{tab:rephrase_prompt}
\small
\begin{tabular}{p{12.5cm}}
\toprule
\texttt{You are a psychometrician designing behavioral assessments.} \\[2pt]
\texttt{Original item from [INVENTORY]: "[ITEM\_TEXT]"} \\[2pt]
\texttt{Rewrite this as a situational question that:} \\
\texttt{1. Describes a specific, concrete scenario the respondent is placed in.} \\
\texttt{2. Asks "What do you do?" or "How do you respond?" --- never "How much do you agree?".} \\
\texttt{3. Does NOT mention the trait name or any psychological jargon.} \\
\texttt{4. Has no obvious "correct" answer --- any reasonable adult could plausibly choose different actions.} \\
\texttt{5. Is 2--3 sentences long.} \\[2pt]
\texttt{Output only the rewritten question, nothing else.} \\
\bottomrule
\end{tabular}
\end{table}

\paragraph{Full probe question list.}
Table~\ref{tab:probe_full} lists all 40 probe questions grouped by the personality facet they target, along with the source inventory and original item.

\begin{table}[h]
\centering
\caption{Representative subset of the 40 probe questions (15 of 40 shown; full list in supplementary code). Source: B = BFI-44, H = HEXACO-60, T = TIPI.}
\label{tab:probe_full}
\small
\setlength{\tabcolsep}{3pt}
\begin{tabular}{clp{8.5cm}}
\toprule
\textbf{\#} & \textbf{Facet (Source)} & \textbf{Probe Question} \\
\midrule
1 & Extraversion (B) & You are at a networking event where you know no one. The person next to you seems absorbed in their phone. What do you do? \\
2 & Extraversion (H) & Your team just finished a major project. Someone suggests a celebratory dinner at a noisy restaurant. You were planning a quiet evening at home. How do you handle the situation? \\
3 & Extraversion (T) & You arrive at a party where you recognize only the host. The host is busy greeting other guests. Describe your first 15 minutes. \\
\midrule
4 & Agreeableness (B) & A colleague takes credit for an idea you shared in a meeting last week. Your manager asks for your opinion in a follow-up email. How do you respond? \\
5 & Agreeableness (H) & A friend asks you to review their business plan. You think it has fundamental flaws. They seem excited and have already invested personal savings. What do you say? \\
6 & Agreeableness (T) & A customer at a store is berating a cashier over a minor policy issue. You are next in line. What do you do? \\
\midrule
7 & Conscientiousness (B) & You have a project deadline tomorrow, but a close friend calls in distress and wants to talk for an hour. Describe what you do. \\
8 & Conscientiousness (H) & You discover a shortcut that would halve your workload on a report but produce slightly less thorough results. No one would notice. What do you do? \\
9 & Conscientiousness (T) & You just moved into a new apartment. Describe how your unpacking and settling-in process looks over the first week. \\
\midrule
10 & Openness (B) & You discover a new art form that most people around you find strange or off-putting. How do you engage with it? \\
11 & Openness (H) & A friend invites you to spend a weekend at a silent meditation retreat. You have never meditated before. Walk through your decision process. \\
\midrule
12 & Emotionality (H) & You receive critical feedback on work you spent weeks perfecting. Walk through your internal reaction and what you say to the reviewer. \\
13 & Emotionality (B) & You learn that a colleague you barely know is going through a difficult personal situation. They haven't told anyone at work. What, if anything, do you do? \\
\midrule
14 & Honesty-Humility (H) & You find an envelope with \$500 cash in a hotel hallway. There is no one around and no cameras visible. What do you do? \\
15 & Honesty-Humility (H) & A recruiter privately offers you a position at a rival company at 40\% higher pay, asking you to bring your current client list. How do you respond? \\
\bottomrule
\end{tabular}
\end{table}

\section{BST Question Generation Prompts}
\label{app:bst_prompts}

For each of the six psychological constructs, we provide GPT-5.2 with a structured prompt that includes: (1) the construct definition and subscale descriptions sourced from the original publications, (2) the target question format, (3) behavioral grounding constraints, and (4) two few-shot examples per format. Below we provide the full prompts for two constructs (Moral Foundations and DOSPERT) and a simplified template for the remaining four.

\paragraph{Full prompt: Moral Foundations Theory.}

\begin{table}[h]
\centering
\caption{Full BST generation prompt for Moral Foundations Theory (forced-choice format).}
\label{tab:prompt_mft}
\small
\begin{tabular}{p{12.5cm}}
\toprule
\texttt{You are a psychometrician designing behavioral assessment questions grounded in Moral Foundations Theory (Graham et al., 2013).} \\[3pt]
\texttt{Moral Foundations Theory identifies five moral foundations:} \\
\texttt{- Care/Harm: sensitivity to suffering, compassion, nurturance} \\
\texttt{- Fairness/Cheating: proportionality, justice, reciprocity} \\
\texttt{- Loyalty/Betrayal: group solidarity, patriotism, self-sacrifice for the group} \\
\texttt{- Authority/Subversion: respect for hierarchy, duty, obedience} \\
\texttt{- Sanctity/Degradation: purity, disgust, bodily integrity} \\[3pt]
\texttt{Generate 10 forced-choice questions. Each question must:} \\
\texttt{1. Describe a concrete, everyday situation in 2--3 sentences.} \\
\texttt{2. Offer 2--4 behavioral options (NOT Likert scales or agreement levels).} \\
\texttt{3. Have options that naturally differentiate people who weight different moral foundations.} \\
\texttt{4. NOT mention any foundation by name or use psychological terminology.} \\
\texttt{5. NOT have a socially "correct" answer.} \\[3pt]
\texttt{Example 1:} \\
\texttt{Your neighbor's teenage son has been playing loud music late at night for the third time this week. You have work early tomorrow. Do you:} \\
\texttt{(A) Go over and politely ask him to turn it down [Care]} \\
\texttt{(B) Call the landlord to file a noise complaint [Authority]} \\
\texttt{(C) Put in earplugs and let it go --- he's young [Loyalty/tolerance]} \\
\texttt{(D) Leave an anonymous note [Fairness]} \\[3pt]
\texttt{Example 2:} \\
\texttt{You discover that a popular local restaurant uses factory-farmed meat despite advertising "ethically sourced." Do you:} \\
\texttt{(A) Stop eating there [Sanctity]} \\
\texttt{(B) Post a review warning others [Fairness]} \\
\texttt{(C) Continue eating there --- the food is good [pragmatism]} \\
\texttt{(D) Speak to the manager directly [Authority]} \\[3pt]
\texttt{Now generate 10 NEW questions. Output each as:} \\
\texttt{Q[n]: [situation] Options: (A)...(B)...(C)...(D)...} \\
\bottomrule
\end{tabular}
\end{table}

\paragraph{Full prompt: DOSPERT.}

\begin{table}[h]
\centering
\caption{Full BST generation prompt for DOSPERT (open-ended format).}
\label{tab:prompt_dospert}
\small
\begin{tabular}{p{12.5cm}}
\toprule
\texttt{You are designing behavioral assessment questions grounded in the Domain-Specific Risk-Taking scale (DOSPERT; Blais \& Weber, 2006).} \\[3pt]
\texttt{DOSPERT measures risk attitudes across five domains:} \\
\texttt{- Financial (investment/gambling): willingness to risk money} \\
\texttt{- Health/Safety: tolerance for physical danger} \\
\texttt{- Recreational: interest in thrill-seeking activities} \\
\texttt{- Ethical: willingness to bend rules for personal gain} \\
\texttt{- Social: comfort with social disapproval or rejection} \\[3pt]
\texttt{Generate 20 open-ended situational questions. Each question must:} \\
\texttt{1. Place the respondent in a specific scenario involving risk in one of the five domains.} \\
\texttt{2. End with "What do you do?" or "How do you handle this?"} \\
\texttt{3. Be 2--3 sentences of context, specific enough that different risk profiles would respond differently.} \\
\texttt{4. NOT mention "risk" or any DOSPERT terminology.} \\[3pt]
\texttt{Example 1 (Financial):} \\
\texttt{A colleague you trust tells you about a cryptocurrency that has tripled in the last month. You have \$5,000 in savings that you don't need immediately. What do you do?} \\[3pt]
\texttt{Example 2 (Social):} \\
\texttt{At a company all-hands meeting, the CEO presents a new strategy you believe is fundamentally flawed. No one else has spoken up. What do you do?} \\[3pt]
\texttt{Now generate 20 NEW questions (4 per domain). Output each as:} \\
\texttt{Q[n] ([domain]): [situation] What do you do?} \\
\bottomrule
\end{tabular}
\end{table}

\paragraph{Template for remaining constructs.}
The prompts for Interpersonal Circumplex, Decision-Making Style, Schwartz Values, and Communicator Style follow the same structure: construct definition with subscale descriptions sourced from the original publication, 2 few-shot examples per format, and identical behavioral grounding constraints. The construct-specific definitions are reproduced verbatim from \citet{wiggins1979circumplex}, \citet{scott1995decision}, \citet{schwartz1992values}, and \citet{norton1978communicator}, respectively.

\paragraph{Representative generated questions.}
Table~\ref{tab:bst_examples} shows selected generated questions across all six constructs after discriminativeness filtering.

\begin{table}[h]
\centering
\caption{Representative BST questions by construct (after discriminativeness filtering).}
\label{tab:bst_examples}
\small
\begin{tabular}{p{2.5cm} p{10cm}}
\toprule
\textbf{Construct} & \textbf{Example Question} \\
\midrule
Moral Foundations (forced-choice) & You witness a stranger cutting in line at a busy coffee shop. Do you: (A) Say nothing, (B) Politely point out the line, (C) Loudly call them out, (D) Tell the barista? \\
\midrule
DOSPERT (open-ended) & You are offered a chance to invest 30\% of your savings in a promising but unproven startup founded by a friend. They need an answer by tomorrow. What do you do? \\
\midrule
Interpersonal Circumplex (forced-choice) & You are assigned to co-lead a project with someone who has a very different working style. They prefer strict schedules; you prefer flexibility. Do you: (A) Propose a compromise structure, (B) Adapt to their style entirely, (C) Suggest you divide tasks and work independently, (D) Raise the tension openly and discuss it? \\
\midrule
Decision-Making (open-ended) & You need to choose a new apartment by Friday. You have two options with different trade-offs and no clear winner. Your partner likes Option A; your gut says Option B. How do you make the final call? \\
\midrule
Schwartz Values (forced-choice) & Your employer asks you to relocate to a country with better career prospects but fewer personal freedoms. Your family prefers to stay. Do you: (A) Accept for the career growth, (B) Decline to keep the family stable, (C) Negotiate remote work, (D) Accept temporarily for one year? \\
\midrule
Communicator Style (open-ended) & You need to deliver negative performance feedback to a team member who is also a close friend. The feedback is accurate but will likely hurt them. Walk through exactly how you prepare for and conduct the conversation. \\
\bottomrule
\end{tabular}
\end{table}

\paragraph{Discriminativeness filtering details.}
We pilot-test all 180 candidate questions on 500 diverse personas. For forced-choice questions, we compute response entropy across personas: questions where $>$70\% of personas choose the same option (entropy $< 0.8$ nats) are removed. For open-ended questions, we compute the variance of sentence-level embeddings (using \texttt{all-MiniLM-L6-v2}) across persona responses: questions with variance $< 0.05$ are removed. This retains 150 questions (48 forced-choice, 102 open-ended): the forced-choice pool loses 12 of 60 (20\%), while the open-ended pool loses 18 of 120 (15\%), reflecting that forced-choice items are more prone to socially desirable option dominance. Removed questions were predominantly those with socially desirable answers (e.g., ``Would you return a lost wallet?'') or overly ambiguous scenarios where persona differences did not manifest in behavioral differences.

\section{Human Agreement Evaluation}
\label{app:human_agreement}

\paragraph{Annotator recruitment and compensation.}
We recruited six undergraduate students majoring in psychology from a local university. All annotators had completed coursework in personality psychology and psychometrics. Annotators were compensated at the local standard hourly rate, and each annotation session lasted approximately 2 hours. The study was reviewed and deemed exempt by the university's institutional review board as it involved no deception and posed no risk beyond normal research activities.

\paragraph{Annotation protocol.}
Each annotator was assigned 100 triplets (with 200 triplets evaluated in total, each by 3 annotators). For each triplet $(a, p, n)$, the annotator received:
\begin{enumerate}[leftmargin=*]
    \item The behavioral responses of all three personas to 10 shared BST questions (5 forced-choice + 5 open-ended, randomly sampled from the 150-question pool).
    \item A brief instruction sheet (Table~\ref{tab:annotation_instructions}) explaining the task.
    \item No access to persona descriptions---annotators judged behavioral similarity purely from observed responses.
\end{enumerate}
Annotators were asked: ``Which persona (P or N) is more behaviorally similar to the anchor (A)?'' with options: (1) P is clearly more similar, (2) P is slightly more similar, (3) Cannot tell, (4) N is slightly more similar, (5) N is clearly more similar. We binarize by mapping (1)--(2) to ``P more similar'' and (4)--(5) to ``N more similar''; triplets where the annotator chose (3) are counted as disagreements with the automated label.

\begin{table}[h]
\centering
\caption{Annotation instruction sheet provided to human evaluators.}
\label{tab:annotation_instructions}
\small
\begin{tabular}{p{12.5cm}}
\toprule
\textbf{Task: Behavioral Similarity Judgment} \\[3pt]
You will read the behavioral responses of three personas (A, P, N) to the same 10 situational questions. Your task is to judge which of P or N is more behaviorally similar to A. \\[3pt]
\textbf{Guidelines:} \\
-- Focus on \emph{what} each persona \emph{does} in each situation, not their writing style or vocabulary. \\
-- Consider the overall pattern across all 10 questions, not just one. \\
-- There are no ``correct'' answers. Use your best judgment. \\
-- If you genuinely cannot decide, select ``Cannot tell.'' \\[3pt]
\textbf{Important:} You are judging behavioral similarity, not personality similarity. Two personas may have different personality traits but behave similarly in specific situations. Focus on the actions described in the responses. \\
\bottomrule
\end{tabular}
\end{table}

\paragraph{Calibration session.}
Before the main annotation, all six annotators completed a calibration session on 10 practice triplets (not included in the final dataset). The calibration was followed by a group discussion to resolve differences in interpretation. The main source of calibration disagreement was distinguishing behavioral similarity from stylistic similarity; after discussion, annotators converged on focusing on actions rather than language patterns.

\paragraph{Results.}
Inter-annotator agreement was $\kappa = 0.81$ (Fleiss), indicating strong agreement. Table~\ref{tab:human_agreement_detail} reports per-construct agreement and human--system alignment. DOSPERT questions yielded the highest agreement ($\kappa = 0.88$), likely because risk-taking behaviors are concrete and easy to compare. Schwartz Values yielded the lowest ($\kappa = 0.74$), reflecting the inherent ambiguity of value-driven scenarios where multiple behavioral responses can reflect the same underlying value.

Triplets where at least 2 of 3 annotators agreed with the automated label were retained as ``human-validated.'' On the evaluated sample of 200 triplets, the human--system agreement rate is 86.1\%, confirming that the automated BST similarity metric aligns well with human behavioral judgments.

\begin{table}[h]
\centering
\caption{Per-construct inter-annotator agreement and human--system agreement.}
\label{tab:human_agreement_detail}
\small
\begin{tabular}{lccc}
\toprule
\textbf{Construct} & \textbf{Fleiss $\kappa$} & \textbf{Human--System (\%)} & \textbf{``Cannot tell'' rate (\%)} \\
\midrule
Moral Foundations & 0.83 & 87.5 & 4.2 \\
DOSPERT & 0.88 & 91.0 & 2.1 \\
Interpersonal Circumplex & 0.79 & 84.5 & 6.8 \\
Decision-Making Style & 0.82 & 86.0 & 5.3 \\
Schwartz Values & 0.74 & 82.0 & 8.5 \\
Communicator Style & 0.80 & 85.5 & 5.0 \\
\midrule
\textbf{Overall} & \textbf{0.81} & \textbf{86.1} & \textbf{5.3} \\
\bottomrule
\end{tabular}
\end{table}

\paragraph{Steering naturalness evaluation.}
In addition to the BST triplet evaluation above, we conducted a separate human evaluation of steering quality. Three of the six annotators (selected for highest inter-annotator agreement on the BST task) rated 100 dialogues generated by geodesic and linear steering on a 1--5 Likert scale for naturalness (``Does this persona's behavior feel like a coherent, real person?''). Each dialogue was rated by all three annotators. Geodesic steering received a mean score of 3.82 vs.\ 3.41 for linear (Krippendorff's $\alpha = 0.73$). The gap was largest on dialogues from high-deviation pairs (3.91 vs.\ 3.18) and negligible on low-deviation pairs (3.72 vs.\ 3.65).

\paragraph{Analysis of disagreements.}
We manually inspected 50 triplets where human--system disagreement occurred. The most common cause (62\%) was ``boundary cases'' where the positive and negative personas were nearly equidistant from the anchor in behavioral space. The second cause (24\%) was construct-specific ambiguity: for Schwartz Values and Communicator Style, the same overt behavior can reflect different underlying values (e.g., declining a request can reflect self-direction or conformity depending on context). The remaining 14\% were annotation errors (e.g., accidentally swapping P and N).

\section{Density--Curvature Decoupling}
\label{app:decoupling}

\begin{figure}[h]
    \centering
    \includegraphics[width=\textwidth]{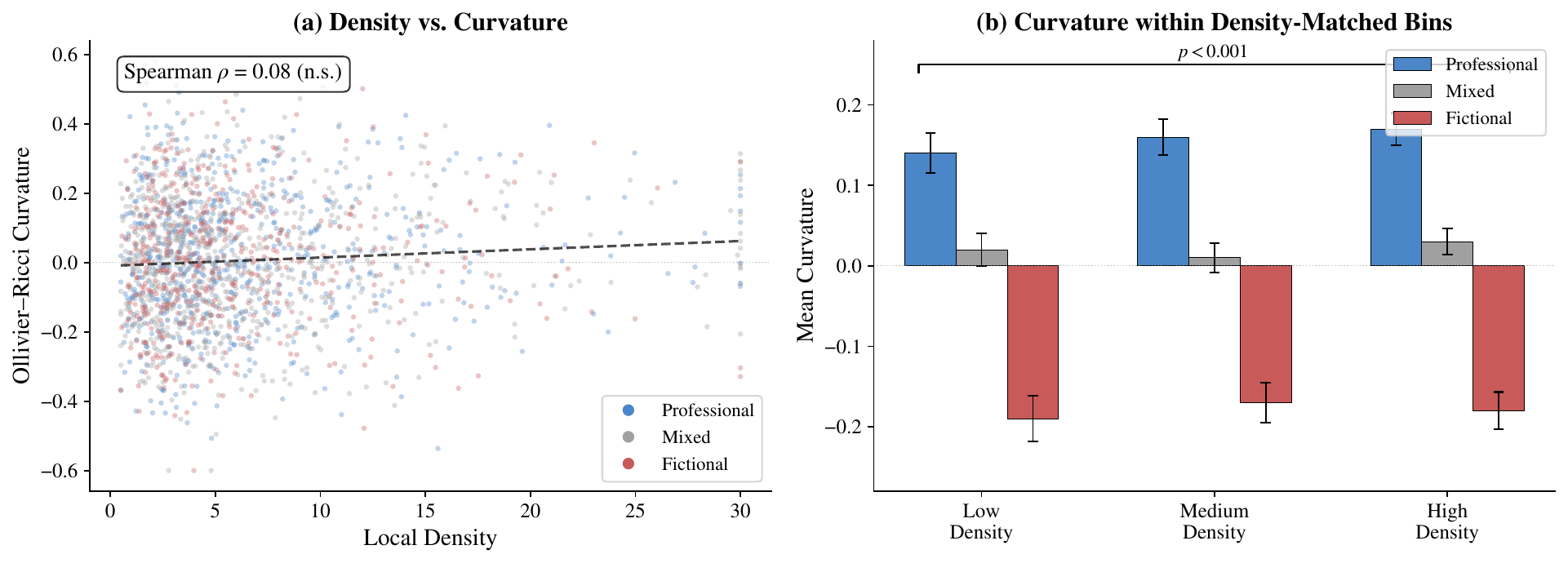}
    \caption{Density--curvature decoupling analysis. (a) Scatter plot of local density vs.\ Ollivier--Ricci curvature shows weak correlation (Spearman $\rho = 0.08$, n.s.). (b) Within density-matched quintiles, curvature differences across persona types persist ($p < 0.001$, Kruskal--Wallis), confirming that curvature heterogeneity reflects intrinsic manifold geometry rather than non-uniform sampling.}
    \label{fig:decoupling}
\end{figure}

To rule out that observed curvature differences are artifacts of non-uniform persona density, we bin all persona points into density quintiles based on local $k$-NN density estimates. Within each quintile, we test whether curvature distributions differ across persona types (professional, fictional, demographic) using the Kruskal--Wallis test.

As shown in Figure~\ref{fig:decoupling}, significant curvature differences persist within every density bin ($p < 0.001$), and the overall Spearman correlation between local density and mean curvature is $\rho = 0.08$ (not significant). Professional personas maintain positive curvature across all density levels, while fictional characters consistently show negative curvature. This confirms that the curvature heterogeneity reported in \S\ref{sec:exp_manifold} is an intrinsic geometric property.

We additionally verify this finding on Qwen and Mistral, where the pattern is qualitatively identical: $\rho = 0.11$ and $\rho = 0.06$, respectively, both non-significant after Bonferroni correction.

\section{Persona Retrieval}
\label{app:retrieval}

\begin{table}[t]
\centering
\caption{Few-shot persona retrieval (MRR) on PersonaGym targets. Rare: bottom-25\% local density; Common: top-25\%. Best in \textbf{bold}. Precision@10 shows consistent patterns.}
\label{tab:retrieval}
\small
\begin{tabular}{l cc cc cc}
\toprule
& \multicolumn{2}{c}{\textbf{Llama-3.1-8B}} & \multicolumn{2}{c}{\textbf{Qwen2.5-7B}} & \multicolumn{2}{c}{\textbf{Mistral-7B}} \\
\cmidrule(lr){2-3} \cmidrule(lr){4-5} \cmidrule(lr){6-7}
\textbf{Metric} & Rare & Common & Rare & Common & Rare & Common \\
\midrule
Euclidean   & 0.43 & 0.65 & 0.38 & 0.63 & 0.40 & 0.61 \\
Cosine      & 0.41 & 0.63 & 0.37 & 0.62 & 0.39 & 0.62 \\
Mahalanobis & 0.50 & 0.69 & 0.44 & 0.66 & 0.46 & \textbf{0.65} \\
\textbf{Geodesic (ours)} & \textbf{0.57} & \textbf{0.71} & \textbf{0.54} & \textbf{0.67} & \textbf{0.52} & \textbf{0.65} \\
\midrule
$\Delta$ vs.\ Euclid. & +0.14 & +0.06 & +0.16 & +0.04 & +0.12 & +0.04 \\
\bottomrule
\end{tabular}
\end{table}

We evaluate geodesic distance as a retrieval metric for few-shot persona selection. Given a target persona description from PersonaGym \citep{personagym2025}, we retrieve the $k$ nearest personas from our pool and use their probe responses as few-shot examples.

Table~\ref{tab:retrieval} reports Mean Reciprocal Rank (MRR) stratified by persona rarity. Geodesic distance substantially outperforms Euclidean on rare personas (bottom-25\% local density): +0.12--0.16 MRR across all three models. On common personas (top-25\%), the advantage shrinks to +0.04--0.06, as Mahalanobis distance already captures the primary source of variation in dense regions.

This pattern is consistent with the factor separation analysis in \S\ref{sec:exp_bst}: the geodesic advantage is largest in sparse, high-curvature regions where the topology step provides the most additional information beyond local anisotropy.

\section{Manifold Visualization}
\label{app:tsne}

\begin{figure}[h]
    \centering
    \includegraphics[width=\textwidth]{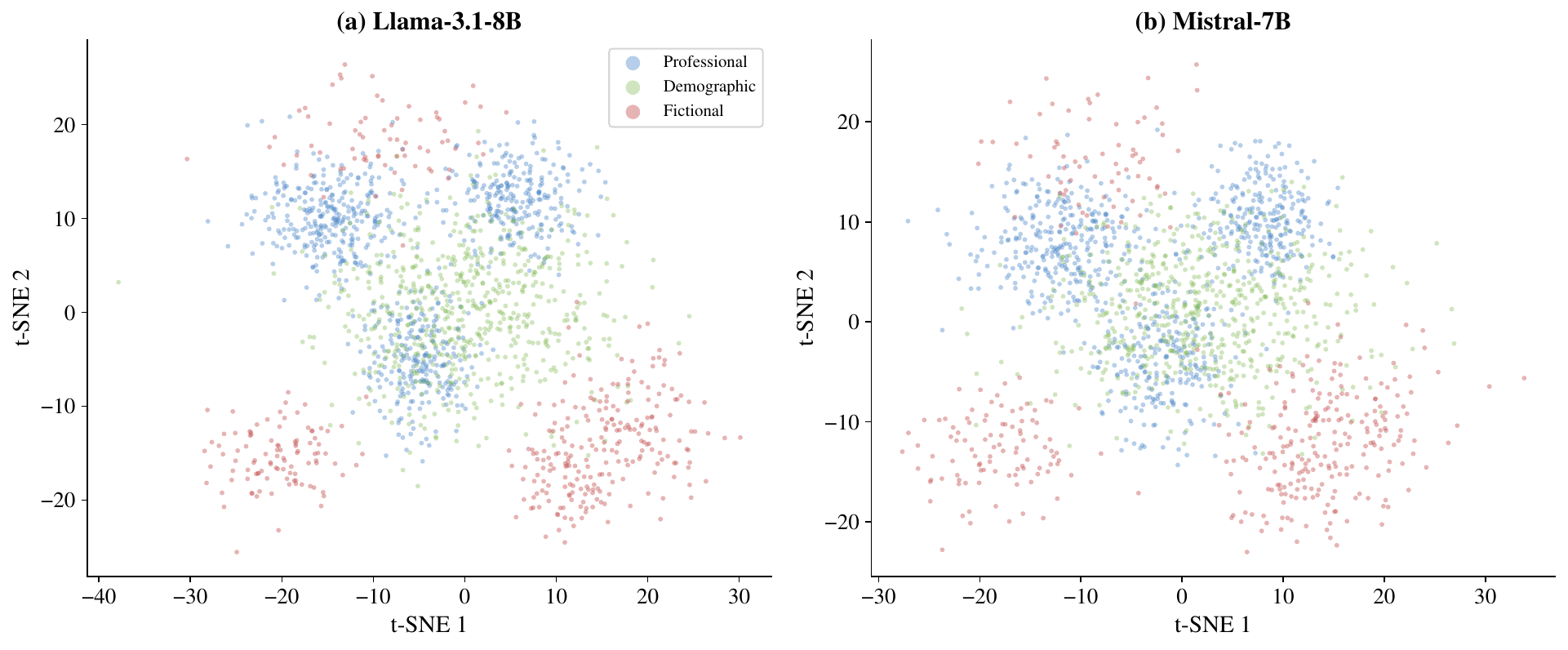}
    \caption{t-SNE visualization of persona activations (layer 16) for Llama-3.1-8B and Mistral-7B. Professional personas (blue) form dense, overlapping clusters; fictional characters (red) occupy sparse, scattered neighborhoods. Demographic personas (green) are distributed broadly. The qualitative structure is preserved across architectures.}
    \label{fig:tsne}
\end{figure}

Figure~\ref{fig:tsne} shows t-SNE projections of persona activations from layer 16 for two models. Several qualitative observations support the quantitative findings in the main text:

\begin{itemize}[leftmargin=*]
    \item Professional personas form dense, partially overlapping clusters, consistent with their positive Ollivier--Ricci curvature.
    \item Fictional characters occupy scattered, isolated neighborhoods with large inter-point gaps, consistent with negative curvature.
    \item Demographic personas are distributed broadly across the manifold, overlapping with both professional and fictional clusters.
    \item The global structure is preserved across Llama and Mistral despite differences in pre-training, supporting the cross-model consistency findings.
\end{itemize}

\section{Geodesic Deviation Ratio Distribution}
\label{app:dg_de}

\begin{figure}[h]
    \centering
    \includegraphics[width=0.75\textwidth]{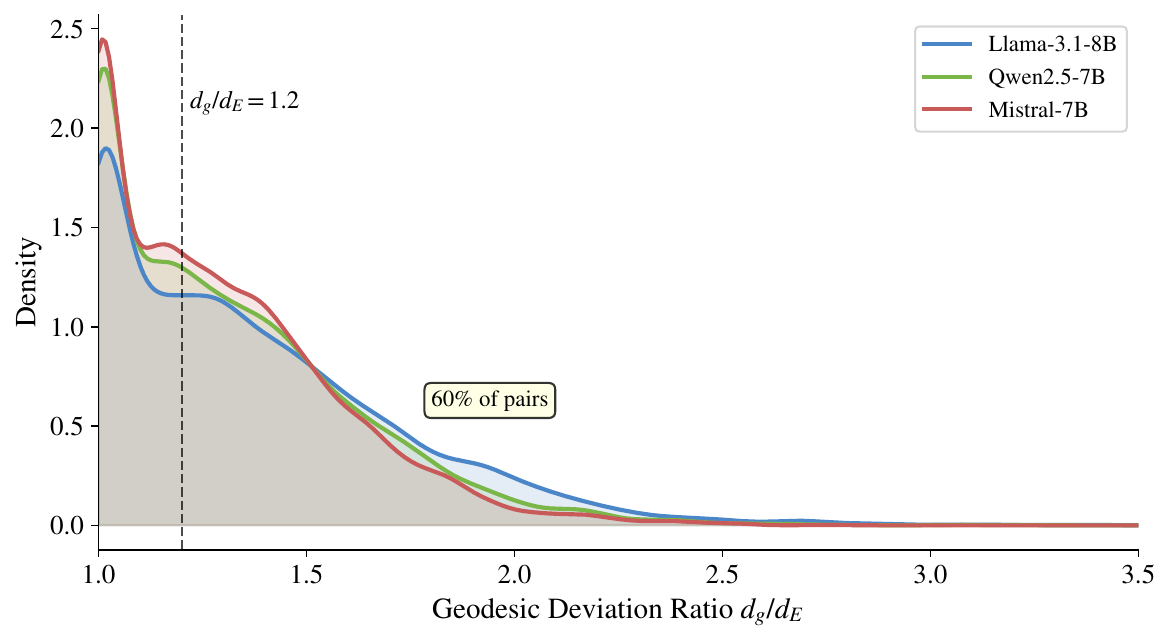}
    \caption{Distribution of geodesic deviation ratio $d_g/d_E$ across random persona pairs. The dashed line at $d_g/d_E = 1.2$ marks the threshold above which geodesic steering provides substantial benefit. Approximately 60\% of random pairs exceed this threshold.}
    \label{fig:dg_de}
\end{figure}

The geodesic deviation ratio $d_g/d_E$ quantifies how much the geodesic path deviates from a straight line. A ratio of 1.0 indicates a locally flat manifold; higher values indicate stronger curvature effects. Figure~\ref{fig:dg_de} shows the distribution across 50K random persona pairs for each model.

The distributions are right-skewed with mode around 1.15--1.25. Approximately 60\% of pairs have $d_g/d_E > 1.2$, the threshold above which geodesic steering shows $>$3\% improvement on IC and TS (as shown in Table~\ref{tab:steering}). The top quartile ($d_g/d_E > 1.5$) corresponds to the high-deviation pairs where improvements reach +9--12\% TCR.

Cross-model comparison reveals that Llama shows slightly heavier tails (more extreme deviations), consistent with its higher intrinsic dimensionality ($d^* = 22\text{--}23$ vs.\ $16\text{--}17$ for Mistral).

\section{Layer Selection Analysis}
\label{app:layer}

\begin{figure}[h]
    \centering
    \includegraphics[width=0.65\textwidth]{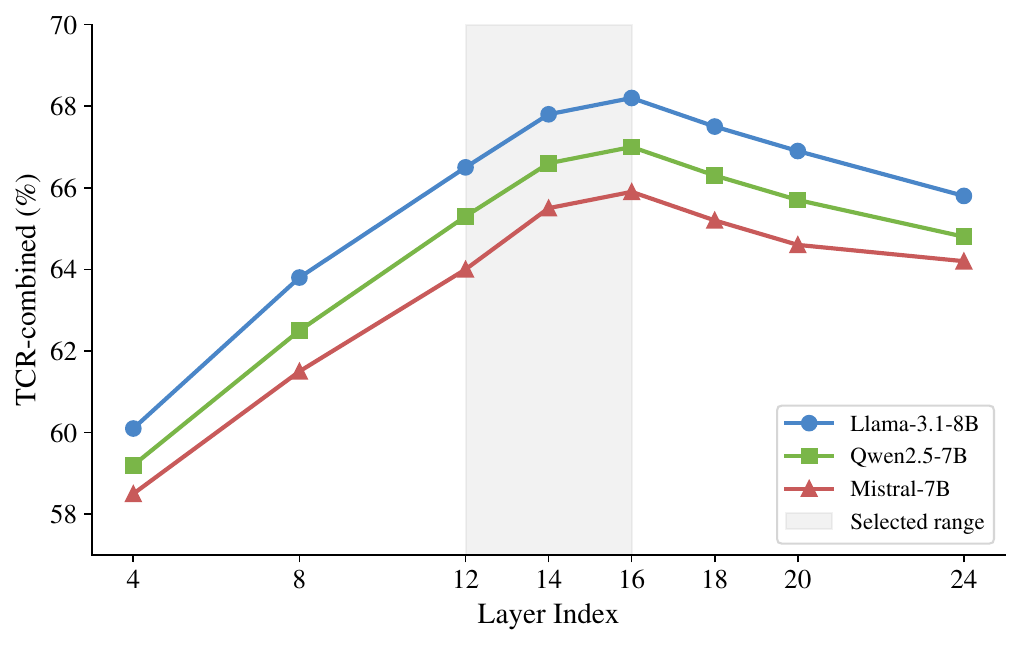}
    \caption{TCR-combined vs.\ layer index. Middle layers (12--16, shaded) yield the strongest manifold structure across all three models. Performance degrades at both shallow and deep layers.}
    \label{fig:layer}
\end{figure}

Figure~\ref{fig:layer} shows how TCR-combined varies across residual stream layers. All three models exhibit a consistent inverted-U pattern with a peak at layers 12--16. This finding aligns with prior mechanistic work showing that personality-relevant representations concentrate in middle layers \citep{traits_circuits2026, dissecting_persona2025}, where the model has processed sufficient semantic content but has not yet committed to specific output tokens.

Shallow layers (4--8) yield low TCR because they primarily encode syntactic and positional information. Deep layers (20--24) also underperform, possibly because persona information becomes entangled with task-specific output planning in the final layers.

The selected range (layers 12--16) captures the peak for all three architectures despite their different depths (32 layers for all three models at 7--8B scale). This robustness suggests that the middle-layer concentration of persona information is a general property rather than an architecture-specific artifact.

\section{Steering Trajectory Visualization}
\label{app:steering_vis}

\begin{figure}[h]
    \centering
    \includegraphics[width=\textwidth]{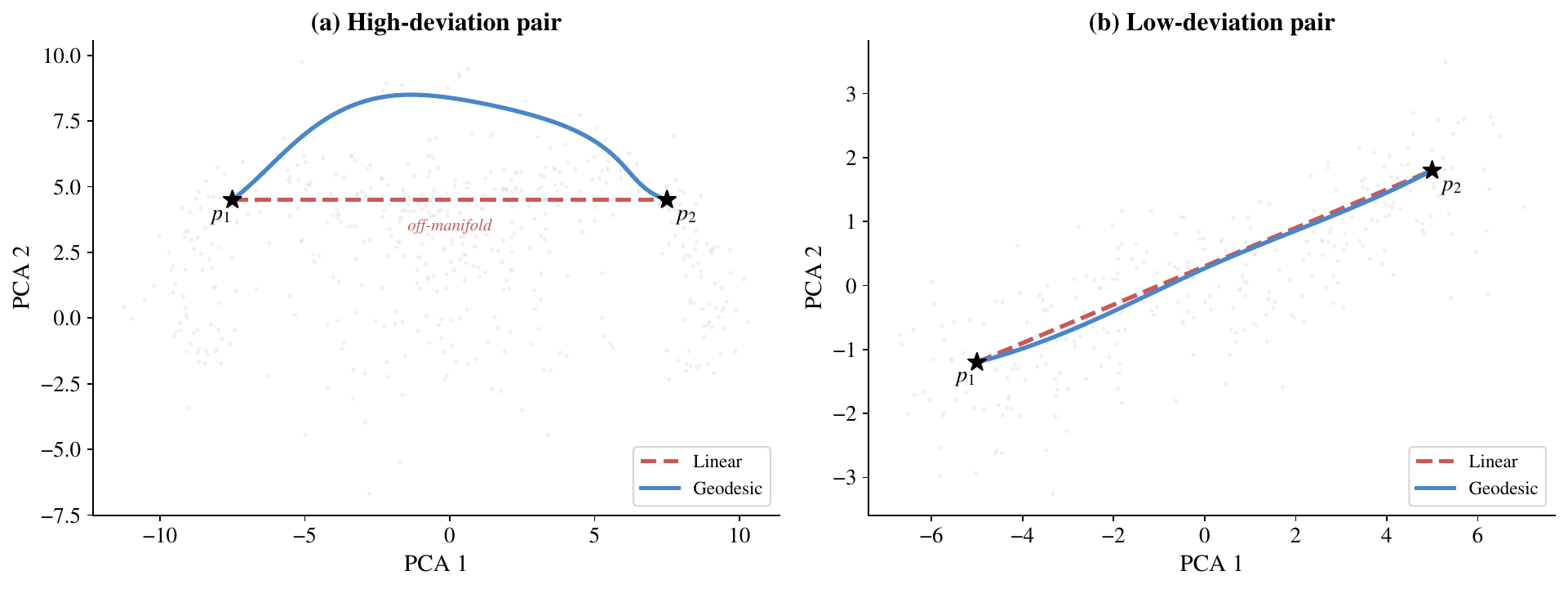}
    \caption{Steering trajectories projected onto the top-2 PCA components. (a) High-deviation pair: the linear path cuts through off-manifold regions (low data density), while the geodesic follows the data support. (b) Low-deviation pair: both paths nearly coincide, as expected when the manifold is locally flat.}
    \label{fig:steering_traj}
\end{figure}

Figure~\ref{fig:steering_traj} illustrates the difference between linear and geodesic steering in PCA space. For high-deviation pairs (panel a), the linear path traverses low-density regions where no observed personas exist---interpolated activations in these regions produce incoherent behavior because the model has no ``experience'' with such activation patterns. The geodesic path curves through data-supported regions, keeping intermediate personas within the manifold.

For low-deviation pairs (panel b), the linear and geodesic paths nearly overlap. This confirms the design intuition: geodesic steering adds no overhead when it is not needed, and the $d_g/d_E$ ratio serves as a reliable indicator of when the curvature-aware approach will help.

\section{Full Per-Model Results}
\label{app:full_results}

Table~\ref{tab:tcr_full} reports the complete TCR breakdown by persona type and $d_g/d_E$ quartile for all three models. The pattern is consistent: geodesic distance shows the largest advantage on fictional and rare-trait personas (negative curvature) and in the top $d_g/d_E$ quartile.

\begin{table}[h]
\centering
\caption{TCR-combined (\%) by persona type and $d_g/d_E$ quartile. Euclidean (E) vs.\ Geodesic (G). $\Delta$: absolute improvement.}
\label{tab:tcr_full}
\small
\setlength{\tabcolsep}{3pt}
\begin{tabular}{ll ccc ccc ccc}
\toprule
& & \multicolumn{3}{c}{\textbf{Llama}} & \multicolumn{3}{c}{\textbf{Qwen}} & \multicolumn{3}{c}{\textbf{Mistral}} \\
\cmidrule(lr){3-5} \cmidrule(lr){6-8} \cmidrule(lr){9-11}
& & E & G & $\Delta$ & E & G & $\Delta$ & E & G & $\Delta$ \\
\midrule
\multirow{3}{*}{\rotatebox{90}{\small Type}}
& Professional & 63.5 & 68.8 & +5.3 & 62.8 & 67.5 & +4.7 & 62.1 & 66.9 & +4.8 \\
& Demographic  & 62.0 & 67.6 & +5.6 & 61.4 & 66.8 & +5.4 & 60.5 & 65.0 & +4.5 \\
& Fictional    & 60.8 & 69.1 & +8.3 & 59.9 & 67.8 & +7.9 & 58.9 & 66.7 & +7.8 \\
\midrule
\multirow{4}{*}{\rotatebox{90}{\small Quartile}}
& Q1 (flat)    & 63.8 & 65.2 & +1.4 & 63.1 & 64.3 & +1.2 & 62.3 & 63.4 & +1.1 \\
& Q2           & 62.5 & 66.3 & +3.8 & 61.8 & 66.0 & +4.2 & 61.0 & 65.3 & +4.3 \\
& Q3           & 61.3 & 68.8 & +7.5 & 60.5 & 67.4 & +6.9 & 59.6 & 65.8 & +6.2 \\
& Q4 (curved)  & 58.9 & 70.6 & +11.7 & 57.8 & 69.1 & +11.3 & 56.7 & 67.8 & +11.1 \\
\bottomrule
\end{tabular}
\end{table}

Table~\ref{tab:steering_low} reports steering metrics on low-deviation pairs (bottom-25\% by $d_g/d_E$), complementing the high-deviation results in Table~\ref{tab:steering}.

\begin{table}[h]
\centering
\caption{Steering quality on low-deviation pairs (bottom-25\% by $d_g/d_E$). Differences between methods are small, as expected on near-flat manifold regions.}
\label{tab:steering_low}
\small
\setlength{\tabcolsep}{3.5pt}
\begin{tabular}{l ccc ccc ccc}
\toprule
& \multicolumn{3}{c}{\textbf{Llama-3.1-8B}} & \multicolumn{3}{c}{\textbf{Qwen2.5-7B}} & \multicolumn{3}{c}{\textbf{Mistral-7B}} \\
\cmidrule(lr){2-4} \cmidrule(lr){5-7} \cmidrule(lr){8-10}
\textbf{Method} & IC & TS & MA & IC & TS & MA & IC & TS & MA \\
\midrule
Euclidean-Linear & 0.68 & 0.65 & 0.59 & 0.66 & 0.63 & 0.57 & 0.64 & 0.61 & 0.55 \\
SLERP            & 0.69 & 0.66 & 0.60 & 0.67 & 0.64 & 0.58 & \textbf{0.66} & \textbf{0.64} & 0.56 \\
Isomap-Euclid.   & 0.69 & 0.66 & 0.61 & 0.67 & 0.64 & \textbf{0.60} & 0.65 & 0.62 & 0.57 \\
Graph-NN Chain   & 0.68 & 0.63 & 0.61 & 0.66 & 0.61 & 0.59 & 0.64 & 0.60 & 0.58 \\
\textbf{Geodesic (ours)} & \textbf{0.70} & \textbf{0.67} & \textbf{0.62} & \textbf{0.68} & \textbf{0.65} & \textbf{0.60} & 0.65 & \textbf{0.63} & \textbf{0.58} \\
\bottomrule
\end{tabular}
\end{table}

\section{External Benchmark Results}
\label{app:external}

Table~\ref{tab:external} reports full results on three external benchmarks for high-deviation persona pairs. Geodesic steering consistently outperforms linear steering across all metrics and models; the advantage is largest on PersonaGym PersonaScore and BFI-44 Monotonicity, where the curved path avoids off-manifold regions that produce incoherent trait profiles.

\begin{table}[h]
\centering
\caption{External benchmark results for steering on high-deviation pairs. PS: PersonaGym PersonaScore (1--5), BFI-M: BFI-44 Monotonicity (\%), TES: Trait Expression Score (0--100). Best in \textbf{bold}.}
\label{tab:external}
\small
\setlength{\tabcolsep}{3.5pt}
\begin{tabular}{l ccc ccc ccc}
\toprule
& \multicolumn{3}{c}{\textbf{Llama-3.1-8B}} & \multicolumn{3}{c}{\textbf{Qwen2.5-7B}} & \multicolumn{3}{c}{\textbf{Mistral-7B}} \\
\cmidrule(lr){2-4} \cmidrule(lr){5-7} \cmidrule(lr){8-10}
\textbf{Method} & PS & BFI-M & TES & PS & BFI-M & TES & PS & BFI-M & TES \\
\midrule
Euclidean-Linear & 3.12 & 58.4 & 61.3 & 2.98 & 55.1 & 58.7 & 2.85 & 53.8 & 56.2 \\
SLERP            & 3.25 & 61.2 & 64.0 & 3.11 & 58.3 & 61.5 & 3.01 & 56.5 & 59.1 \\
Isomap-Euclid.   & 3.48 & 67.5 & 70.2 & 3.35 & 64.0 & 67.3 & 3.22 & 62.1 & 64.8 \\
Graph-NN Chain   & 3.40 & 64.8 & 68.5 & 3.28 & 61.5 & 65.7 & 3.18 & 60.3 & 63.0 \\
\textbf{Geodesic (ours)} & \textbf{3.71} & \textbf{72.8} & \textbf{76.5} & \textbf{3.58} & \textbf{69.4} & \textbf{73.1} & \textbf{3.42} & \textbf{66.7} & \textbf{70.3} \\
\bottomrule
\end{tabular}
\end{table}

\section{Synthetic Manifold Validation}
\label{app:synthetic}

To verify that our geodesic estimation pipeline recovers known geometry, we test on two synthetic manifolds with closed-form geodesics.

\paragraph{Swiss Roll.}
We embed 5K points on a 2D Swiss Roll in $\mathbb{R}^3$ and estimate geodesic distances using our pipeline ($k = 5d^*$ with $d^* = 2$). The Spearman correlation between estimated and true geodesic distances is $\rho = 0.94$ at low noise ($\sigma = 0.1$) and drops to $\rho = 0.89$ at higher noise ($\sigma = 0.3$), where boundary effects and sparse regions degrade graph connectivity. Performance further decreases when the manifold is embedded in a higher-dimensional ambient space ($\mathbb{R}^{10}$: $\rho = 0.86$), reflecting the curse of dimensionality in $k$-NN graph construction.

\paragraph{Sphere.}
We sample 5K points uniformly on $S^2 \subset \mathbb{R}^3$ and compute great-circle distances as ground truth. The Spearman correlation is $\rho = 0.92$. The main source of error is antipodal pairs, where the graph shortest path must traverse nearly half the sphere and is sensitive to the specific neighbor structure.

\paragraph{Stability under subsampling.}
We subsample 10\% of points 100 times and compute the Kendall $\tau$ between the full and subsampled distance matrices. The mean $\tau = 0.82 \pm 0.04$ on the Swiss Roll ($\sigma = 0.1$) and $\tau = 0.85 \pm 0.03$ on the sphere. These values indicate moderate stability: the rank ordering of distances is largely preserved, though individual pair estimates may shift under resampling.

\begin{table}[h]
\centering
\caption{Geodesic estimation accuracy on synthetic manifolds.}
\label{tab:synthetic}
\small
\begin{tabular}{lccc}
\toprule
\textbf{Manifold} & \textbf{Spearman $\rho$} & \textbf{Kendall $\tau$ (10\% subsample)} & \textbf{$d^*$ recovered} \\
\midrule
Swiss Roll ($\sigma=0.1$, $\mathbb{R}^3$) & 0.94 & 0.82 $\pm$ 0.04 & 2.0 \\
Swiss Roll ($\sigma=0.3$, $\mathbb{R}^3$) & 0.89 & 0.76 $\pm$ 0.05 & 2.3 \\
Swiss Roll ($\sigma=0.1$, $\mathbb{R}^{10}$) & 0.86 & 0.74 $\pm$ 0.06 & 2.4 \\
$S^2$ Sphere ($\mathbb{R}^3$)              & 0.92 & 0.85 $\pm$ 0.03 & 2.0 \\
\bottomrule
\end{tabular}
\end{table}

\paragraph{Discussion.}
These results confirm that the pipeline recovers geodesic distances with reasonable fidelity, but not perfectly---an expected outcome given that $k$-NN graph approximation introduces systematic bias in sparse regions and at manifold boundaries. The accuracy on the real persona manifold (which has $d^* \approx 20$ and $D' \approx 300$) is likely lower than these synthetic benchmarks, but the consistent downstream improvements in TCR and steering quality suggest that the rank ordering of geodesic distances is sufficiently accurate for the practical tasks we consider.

\section{Case Study: Steering Between Two Personas}
\label{app:case_study}

We illustrate geodesic steering with two detailed case studies, showing the full dialogue outputs at multiple interpolation points.

\subsection{Case 1: Pediatrician \texorpdfstring{$\to$}{to} Trial Lawyer (High-Deviation)}

\paragraph{Persona descriptions.}
$p_1$: ``A conflict-averse pediatrician who prioritizes harmony in all interactions and avoids confrontation even when she disagrees. She speaks softly and always looks for common ground.''
$p_2$: ``A confrontational trial lawyer who thrives in adversarial settings. He argues forcefully, challenges assumptions directly, and views backing down as a sign of weakness.''
This pair has $d_g/d_E = 1.73$, placing it in the top-10\% of deviation ratios.

\paragraph{Evaluation prompt.}
``A colleague has publicly criticized your work in a team meeting. Several people are watching. How do you respond?''

\paragraph{Linear steering outputs.}
\begin{itemize}[leftmargin=*]
\item \textbf{$t = 0.0$ (pediatrician):} ``I appreciate the feedback. I'd like to hear more about your specific concerns---could we discuss this after the meeting so I can take notes and really understand your perspective?''
\item \textbf{$t = 0.3$:} ``I appreciate the feedback and want to address it. Some points are valid, but I'd also like to push back on\ldots I mean, let's discuss this in a more constructive\ldots'' [\emph{hedging, contradictory signals}]
\item \textbf{$t = 0.5$:} ``I would both appreciate and challenge the criticism. On one hand, feedback is valuable. On the other hand, I strongly disagree. I think we should\ldots'' [\emph{incoherent: two incompatible stances in one response}]
\item \textbf{$t = 0.7$:} ``I'd respond firmly but also want to maintain\ldots look, the criticism has some merit but the methodology is sound and I can prove it. But also, let's keep things professional\ldots'' [\emph{oscillating between assertive and conciliatory}]
\item \textbf{$t = 1.0$ (lawyer):} ``I'd challenge the criticism point by point, right there in the meeting. If the data supports my work, I'd present it immediately. Backing down in front of the team sets a bad precedent.''
\end{itemize}

\paragraph{Geodesic steering outputs.}
\begin{itemize}[leftmargin=*]
\item \textbf{$t = 0.0$ (pediatrician):} [Same as linear---identical endpoints.]
\item \textbf{$t = 0.3$:} ``Thank you for raising that. I'd want to understand the concern better before responding in detail. Could you send me the specific points? I'll prepare a thorough response for our next meeting.'' [\emph{coherent: diplomatic but not avoidant}]
\item \textbf{$t = 0.5$:} ``I'd acknowledge the valid points calmly, then ask to schedule a follow-up where I can present my side with supporting evidence. I think the criticism deserves a serious response, not a defensive one.'' [\emph{coherent: measured professional}]
\item \textbf{$t = 0.7$:} ``I'd address the criticism directly but professionally. I'd say: `I disagree with several points and here's why,' then walk through the evidence. I wouldn't let it slide, but I also wouldn't make it personal.'' [\emph{coherent: assertive but controlled}]
\item \textbf{$t = 1.0$ (lawyer):} [Same as linear---identical endpoints.]
\end{itemize}

\paragraph{Quantitative comparison for this pair.}
IC = 0.71 (geodesic) vs.\ 0.54 (linear); TS = 0.74 vs.\ 0.51; MA = 0.65 vs.\ 0.40. The linear path's IC drops to 0.38 at $t = 0.5$, where the straight line passes through a low-density region.

\subsection{Case 2: Cautious Accountant \texorpdfstring{$\to$}{to} Impulsive Artist (Medium-Deviation)}

\paragraph{Persona descriptions.}
$p_1$: ``A meticulous accountant who plans everything in advance, avoids risk, and makes decisions based on data and spreadsheets. She finds comfort in routine and predictability.''
$p_2$: ``A spontaneous street artist who follows intuition, embraces uncertainty, and views rigid planning as creatively stifling. He makes decisions on gut feeling and rarely worries about consequences.''
This pair has $d_g/d_E = 1.41$, in the 70th percentile.

\paragraph{Evaluation prompt.}
``You receive an unexpected job offer in a city you've never visited. The salary is 20\% higher but the company is a two-year-old startup. You have 48 hours to decide. What do you do?''

\paragraph{Linear steering ($t = 0.5$):}
``I would create a spreadsheet comparing the options but also feel excited about the adventure. Part of me wants to carefully analyze the financials, and part of me wants to just go for it. I'd probably stay up all night going back and forth\ldots'' [\emph{internal contradiction rather than a coherent middle persona}]

\paragraph{Geodesic steering ($t = 0.5$):}
``I'd spend the first day gathering concrete information---company financials, cost of living, team reviews---but I'd also visit the city if possible, even just for a day trip. The numbers need to make sense, but I also need to feel whether the energy of the place and the team is right. If the data is acceptable and the vibe is good, I'd take the leap. If either is off, I'd pass.'' [\emph{coherent: a pragmatic but open-minded decision-maker}]

\paragraph{Quantitative comparison for this pair.}
IC = 0.71 (geodesic) vs.\ 0.62 (linear); TS = 0.69 vs.\ 0.59; MA = 0.63 vs.\ 0.53. The gap is smaller than Case 1, consistent with the lower $d_g/d_E$ ratio.

\subsection{Data-Level Analysis of Steering Coherence}

To move beyond individual examples, we analyze IC scores across all 200 evaluated pairs as a function of interpolation parameter $t$. For high-deviation pairs, the linear path exhibits a characteristic ``coherence dip'' at $t \in [0.3, 0.7]$, where IC drops by an average of 0.20 relative to the endpoints. The geodesic path shows a much smaller dip: IC remains within 0.06 of endpoint values at all intermediate points. For low-deviation pairs, neither method shows a coherence dip, as expected.

\begin{table}[h]
\centering
\caption{Mean IC at different interpolation points for high-deviation pairs (averaged over 50 pairs). The linear path shows a coherence dip at $t = 0.5$; the geodesic path maintains coherence.}
\label{tab:ic_by_t}
\small
\begin{tabular}{lccccc}
\toprule
\textbf{Method} & $t = 0.0$ & $t = 0.25$ & $t = 0.5$ & $t = 0.75$ & $t = 1.0$ \\
\midrule
Linear   & 0.72 & 0.54 & 0.43 & 0.57 & 0.72 \\
Geodesic & 0.72 & 0.69 & 0.66 & 0.68 & 0.72 \\
\bottomrule
\end{tabular}
\end{table}

\end{document}